\documentclass[lettersize,journal]{IEEEtran}
\usepackage{amsmath,amsfonts}
\usepackage{algorithmic}
\usepackage{array}
\usepackage[caption=false,font=normalsize,labelfont=sf,textfont=sf]{subfig}

\usepackage{textcomp}
\usepackage{phonetic}
\usepackage{tipa}

\usepackage{stfloats}
\usepackage{url}
\usepackage{verbatim}
\usepackage{graphicx}
\usepackage{balance}
\usepackage{amsmath}
\usepackage{cite}
\usepackage{bbm}
\usepackage{multirow}
\usepackage{hyperref} 
\usepackage{tabularx}
\hypersetup{
    colorlinks=true,   
    linkcolor=blue,    
    citecolor=blue,     
    urlcolor=magenta   
}
\usepackage{breqn}

\usepackage{phonetic}
\newcommand{\ipa}[1]{\textipa{#1}} 
\usepackage{makecell}

\begin{document}

\title{
UINO-FSS: Unifying Representation Learning and Few-shot Segmentation via Hierarchical Distillation and Mamba-HyperCorrelation
}

\author{Wei Zhuo, Zhiyue Tang, Wufeng Xue\textsuperscript{\dag},~\IEEEmembership{Member, IEEE}, Hao Ding, Junkai Ji, Linlin Shen\textsuperscript{\dag},~\IEEEmembership{Senior Member, IEEE}
\thanks{\textsuperscript{\dag} Wufeng Xue and Linlin Shen are the corresponding authors.}
\thanks{Wei Zhuo, Zhiyue Tang, Hao Ding, and Junkai Ji are with the School of Artificial Intelligence and the National Engineering Laboratory of Big Data System Computing Technology, Shenzhen University, Shenzhen 518060, China. Wei Zhuo is also with the Guangdong Provincial Key Laboratory of Intelligent Information Processing, China (e-mail: weizhuo@szu.edu.cn, tangzhiyue2023@email.szu.edu.cn, 2400671002@mails.szu.edu.cn, jijunkai@szu.edu.cn).}
\thanks{Wufeng Xue is with National-Regional Key Technology Engineering Laboratory for Medical Ultrasound, School of Biomedical Engineering, Shenzhen University Medical School, Shenzhen University, Shenzhen 518060,
China (e-mail: xuewf@szu.edu.cn).} 
\thanks{ Linlin Shen is with the School of Artificial Intelligence and the National Engineering Laboratory of Big Data System Computing Technology, Shenzhen University, Shenzhen 518060, China, and also with the Department of Computer Science, University of Nottingham Ningbo
China, Ningbo 315100, China (e-mail: llshen@szu.edu.cn).}
\thanks{This project is supported by National Natural Science Foundation of China under Grant 62306183, 62471313, Guangdong Natural Science Foundation under Grant 2024A1515010194, 2024A1515030143, Shenzhen Natural Science Foundation under Grant JCYJ20240813141807010, and the Intelligent Computing Center of Shenzhen University.}
}

\markboth{Journal of \LaTeX\ Class Files,~Vol.~14, No.~8, August~2021}%
{Shell \MakeLowercase{\textit{et al.}}: A Sample Article Using IEEEtran.cls for IEEE Journals}


\maketitle

\begin{abstract}
Few-shot semantic segmentation has attracted growing interest for its ability to generalize to novel object categories using only a few annotated samples. To address data scarcity, recent methods incorporate multiple foundation models to improve feature transferability and segmentation performance. However, they often rely on dual-branch architectures that combine pre-trained encoders to leverage complementary strengths, a design that limits flexibility and efficiency. This raises a fundamental question: \textit{``can we build a unified model that integrates knowledge from different foundation architectures?"} Achieving this is, however, challenging due to the misalignment between class-agnostic segmentation capabilities and fine-grained discriminative representations. To this end, we present UINO-FSS (pronounced  /\ipa{ju:"aIn@U}/), a novel framework built on the key observation that early-stage DINOv2 features exhibit distribution consistency with SAM's output embeddings. This consistency enables the integration of both models' knowledge into a single-encoder architecture via coarse-to-fine multimodal distillation. In particular, 
our segmenter consists of three core components: a bottleneck adapter for embedding alignment, a meta-visual prompt generator that leverages dense similarity volumes and semantic embeddings, and a mask decoder. Using hierarchical cross-model distillation, we effectively transfer SAM's knowledge into the segmenter, further enhanced by Mamba-based 4D correlation mining on support-query pairs. 
Extensive experiments show that UINO-FSS achieves new state-of-the-art results on COCO-20$^i$ under the 1-shot setting, with an mIoU of 64.5\% (+2.2\%), while also delivering competitive performance on PASCAL-5$^i$.
\end{abstract}

\begin{figure}[h!]
\centering
\includegraphics[width=0.5\textwidth]{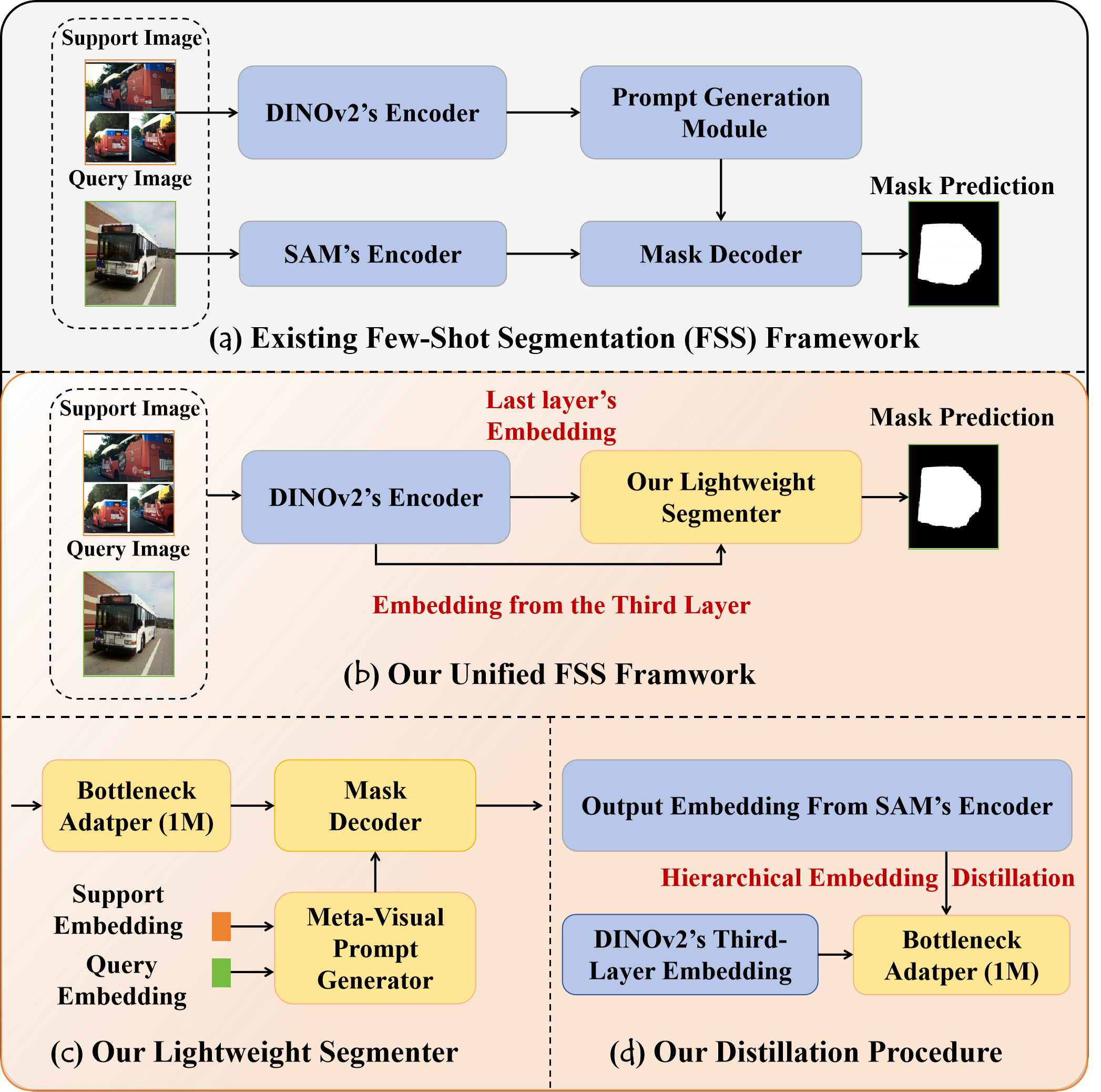}
\caption{\textbf{Introduction of our framework.} Our framework (b) is unified and efficient compared to the existing dual-modal architectures~\cite{liu2024matcher,zhang2025bridge,sun2024vrp} shown in (a). Our lightweight segmenter (c) only contains 5.6M parameters, but gains knowledge from the powerful SAM via the procedure in (d). }
\label{fig:introfig}
\vspace{-8pt}
\end{figure}


\section{Introduction}
\IEEEPARstart{S}{emantic} segmentation, which performs pixel-wise classification, has achieved great success in recent years. However, these methods usually require extensive training on large-scale, densely annotated datasets. To mitigate the burden of manual annotation, Few-shot Semantic Segmentation (FSS) has attracted increasing attention. Unlike traditional semantic segmentation, FSS aims to build a generalizable framework that can segment novel classes using only a few labeled examples.

Inspired by \cite{snell2017prototypical}, early works~\cite{shaban2017one,wang2019panet} employ meta-learning to learn generalized matching models through extensive meta-tasks constructed from support-query pairs. Such a design, which closely mimics the few-shot inference, enables the learned model to be directly applied to segment novel classes without finetuning during inference. To enable effective support-query matching, prototype-based methods~\cite{liu2020part,yang2020prototype,latent-mining, zhang2021self,zhang2020sg, gao2022drnet,chen2024dual} simply match the class prototype vector with each query pixel, while approaches based on embedding aggregation~\cite{zhang2021few, iqbal2022msanet,peng2023hierarchical,min2021hypercorrelation, hong2022cost} extensively fuse the support and query feature maps for subsequent segmentation. 
Here, for embedding aggregation, convolutional hypercorrelation~\cite{min2021hypercorrelation} and diverse attention-based matching~\cite{zhang2021few,iqbal2022msanet,peng2023hierarchical, xu2023self} are applied.
Despite these efforts, FSS still suffers heavily from the inherent limitation of data scarcity and faces challenges in achieving stable and distinguished embeddings of novel classes.

\begin{figure*}[htbp]
\centering
\includegraphics[width=0.95\textwidth]{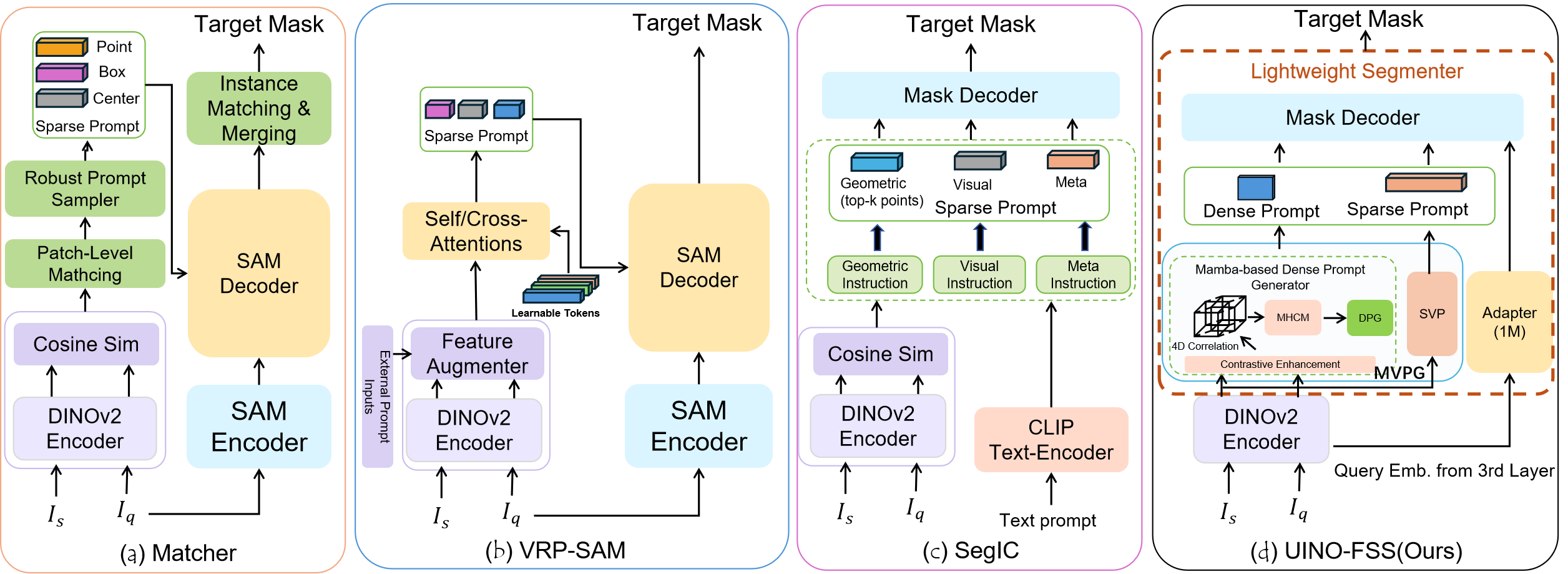}
\caption{\textbf{Framework comparison with prior FSS works.} Existing SAM-based FSS methods, such as Matcher~\cite{liu2024matcher} and VRP-SAM~\cite{sun2024vrp}, typically adopt a dual-encoder design, using SAM's encoder and mask decoder alongside a separate DINOv2 extractor for support-query feature extraction. SegIC adopts a single-encoder framework but requires CLIP for meta prompts. In contrast, UINO-FSS employs a single DINOv2 encoder with a lightweight bottleneck adapter, extracting features only once without any additional extractor. For prompt generation, prior methods rely on sparse prompts (points/boxes) derived from 2D distance maps between support prototypes and query features. UINO-FSS provides a unified framework with Meta-Visual Prompt Generator (MVPG) that eliminates sparse points and boxes entirely, relying instead on dense prompts primarily. Specifically, we model 4D correlations between support feature map and the query correspondence via contrastive enhancement and a Mamba-HyperCorrelation Module (MHCM) followed by Dense Prompt Generation (DPG), while introducing semantic-aware sparse prompts that replace explicit point/box inputs.}
\label{fig:compare-framework}
\vspace{-8pt}
\end{figure*}

In recent years,  FSS methods based on foundation models~\cite{caron2021emerging,oquab2024dinov2,radford2021learning,kirillov2023segment} have pushed few-shot segmentation further to new SOTA accuracy by exploiting visual representations learned from large-scale data. Among them, the DINO family~\cite{caron2021emerging,oquab2024dinov2} stands out for preserving fine-grained spatial correlations, a property that makes the features almost tailor-made for dense matching in FSS. DINOv2, however, is trained with pure self-supervision without a segmentation head; its frozen backbone alone is insufficient for producing robust masks. SAM~\cite{kirillov2023segment}, on the other hand, couples a ViT image encoder with a lightweight, class-agnostic mask decoder trained on 1B+ annotated pixels, which has inspired a branch of SAM-augmented FSS pipelines~\cite{liu2024matcher,he2024apseg,zhang2023personalize,zhang2025bridge}. However, SAM itself also bears limitations in extracting fine-grained embeddings for categorization. To leverage the advantages of both foundation models, recent methods confront the intrinsic misalignment between SAM's task-specific embeddings and DINOv2's generic features, thereby resorting to a dual-branch design~\cite{liu2024matcher,zhang2025bridge,sun2024vrp}. 
That is, current state-of-the-art FSS frameworks commonly adopt SAM's encoder-decoder for image embedding and mask decoding, while using DINOv2's encoder to extract fine-grained features from both support and query images for matching and sparse prompts (points/boxes) generation.

Despite the achievement, existing FSS methods based on foundation models face the following challenges: 
1) Hybrid frameworks that run two separate parallel branches, as shown in Fig.~\ref{fig:introfig}(a) and Fig.~\ref{fig:compare-framework}(a-b), would inevitably lead to high computational and memory costs, resulting in low efficiency and limited development in resource-constrained scenarios; 
2)
Beyond computation, SAM-based FSS methods still primarily rely on geometric sparse prompts of points or boxes (see Fig.~\ref{fig:compare-framework}), which could be sensitive to the point prompt selection. To overcome these issues, Matcher\cite{liu2024matcher} designs point sampler with bi-directional prompt selection to prune noisy prompts. SegIC~\cite{meng2024segic} introduces top-k points with visual and text prompts. Later work~\cite{zhang2025bridge} constructs a graph for segment filtering and merging processes, which, however, still relies on the basic segmentation capability of SAM. Despite recent advances such as VRP-SAM, which employs learnable query vectors as sparse prompts, its prompt vectors lack spatial cues.   3) ViT self-attention lets each patch aggregate global context, yet co-occurring objects in the support image would also contaminate the support features due to the~\textit{embedding coupling}, where embeddings of one object incorporate semantics from other co-occurring objects.

To this end, we propose UINO-FSS (pronounced  /\ipa{ju:"aIn@U}/),
a novel, efficient framework that \textit{\textbf{U}}nifies D\textit{\textbf{INO}}v2-based representation learning and \textit{\textbf{F}}ew-\textit{\textbf{S}}hot \textit{\textbf{S}}eg-mentation. As discussed above, to leverage the complementary advantages from both SAM and DINOv2, current frameworks typically adopt a dual-branch architecture combining SAM's encoder-decoder for segmentation and DINOv2 for prompt generation. However, through our observation on self-similarity maps and the quantitative results (see \textit{Preliminary Sec.}), we found that DINOv2's early-stage features are highly consistent with features from SAM's last layer. This finding indicates that the two feature spaces are sufficiently compatible to support a unified single-stream pipeline without maintaining redundant encoders. Therefore, we design a unified framework consisting of a DINOv2 encoder and a lightweight segmenter (as shown in Fig.~\ref{fig:introfig}(b–d)). Our segmenter draws on architectural insights from SAM's decoder while achieving superior DINOv2 compatibility via adapter-distillation for feature alignment, and integrates a Meta-Visual Prompt Generator (MVPG) that automatically generates prompts from support-query metadata. The MVPG entirely eliminates handcrafted point/box extraction through a Mamba-based dense prompt generator and a semantic-aware visual prompt generator.

Specifically, we introduce a lightweight Bottleneck Adapter (BA) ($\sim$1M parameters) atop DINOv2's early-stage layers to align its features with SAM's output space, thereby adapting them to encode holistic objects for segmentation. We enforce precise cross-model alignment through stage-wise hierarchical knowledge distillation across multiple resolutions. Formally, the distillation objective minimizes the $\ell_2$ distance between patch-level features at corresponding spatial locations in both networks, progressively aligning local feature distributions and promoting global consistency. Compared to prior DINOv2-based FSS methods~\cite{meng2024segic}, this knowledge consolidation achieves significantly better performance with only limited parameters.

For automatic prompting, prior methods extract points/boxes based on 2D support-query similarity, while others directly leverage the 2D similarity map between support prototypes and the query feature map. However, we argue that since the 2D support prototype vectors lack spatial cues, the resulting 2D distance map inevitably misses fine-grained spatial correspondence. 
To boost segmentation quality, we introduce the MVPG with a novel Mamba-based dense prompt generator, enabling foreground-background contrastive enhancement and Mamba-based hypercorrelation processing. Comparing patches between the query and support foreground, it leads to a 4D correlation volume, named hypercorrelation. Beyond~\cite{min2021hypercorrelation} that decouples the hypercorrelation into 2D slices along the support and query dimensions, we introduce an efficient Mamba-HyperCorrelation Module (MHCM) that processes the full volumetric similarities as an integrated 4D structure, enabling true high-order correlation modeling and fine-grained matching. The 4D correlation volume is hierarchically encoded via stacked Hierarchical Global Modeling Blocks (HGMB), which alternate between convolutions and Visual State Space (VSS) layers to progressively capture both local and global dependencies, thereby improving robustness to noise and matching accuracy.
Additionally, we introduce a Contrastive Enhancement (CE) process prior to MHCM. In particular, by simply subtracting the similarities between support background and the query from the hypercorrelation, the subsequent MHCM can acutely capture the embedding entanglement cues, thus suppressing erroneous correspondence for more accurate segmentation. 

To enable effective training, we introduce a hybrid strategy combining coarse-to-fine knowledge distillation for embedding alignment with meta-learning for few-shot semantic segmentation (FSS). Overall, we summarize our key contributions as follows:

\begin{itemize}
\item We present a pivotal observation (Sec. \ref{sec:Preliminary}) that serves as the foundation for our unified framework, UINO-FSS, which adopts a single-branch architecture consisting solely of a DINOv2 encoder and a lightweight segmenter. The segmenter that enables mask decoding with automatic prompting, contains only 5.6M parameters (about 5\% of SAM's) while maintaining strong performance.
\item We introduce the Meta-Visual Prompt Generator (MVPG) with the novel Mamba-HyperCorrelation module and contrastive enhancement. That processes 4D correlation with reduced embedding coupling.
\item A family of compact UINO-FSS models on DINOv2 variations (DINOv2-Small, Base, Large) is trained and validated, proving the generalization of the framework. 
Our proposed method achieves State-of-the-Art (SOTA) performance on the COCO-20$^i$ benchmark~\cite{nguyen2019feature}, with a dramatic improvement of 2.2\% under the challenging 1-shot setting, while delivering competitive performance on PASCAL-5$^i$~\cite{shaban2017one}, with only 0.4\% lower mIoU but substantially fewer total parameters (92.5M vs. 132.6M) compared to the previous best model (FSSAM~\cite{xu2025unlocking}).
\end{itemize}

\section{Related Work}
\subsection{Semantic Segmentation}
Semantic segmentation, which involves assigning a semantic label to each pixel, constitutes a fundamental task in computer vision.  A key challenge in this area lies in the effective integration of global and local feature encoding. 
Early architectures such as UNet \cite{ronneberger2015u, badrinarayanan2017segnet} preserve local spatial details through skip connections, while subsequent methods used dilated convolutions \cite{yu2015multi,chen2017rethinking,chen2018encoder} and multi-scale modules \cite{zhao2017pyramid,sun2019high,qin2024pyramid} to capture broader contextual information and better global semantics. Techniques such as \cite{fu2019dual} and \cite{huang2019ccnet} insert dual-attention and cross-attention, respectively, into convolutional backbones, further improving contextual reasoning. 
More recently, Vision Transformer (ViT)-based frameworks \cite{cheng2022masked, xie2021segformer, wen2024rethinking} combine the ViT encoder with a mask decoder, leading to a unified architecture. Unlike traditional methods that predict pixel labels independently, self-attention in the mask decoder allows object queries to perceive the others, facilitating global reasoning. Following this trend, SAM \cite{kirillov2023segment} incorporates prompt encoding and achieves strong class-agnostic segmentation via large-scale pretraining. While SAM excels in global boundary detection at the cost of local details, DINOv2 provides rich local features but lacks a powerful decoder, underscoring the ongoing efforts of unifying global and local cues.

\subsection{Few-shot Semantic Segmentation}
Instead of supervising on massive annotations, few-shot semantic segmentation aims to segment novel semantics using only a limited number of annotated samples. In this field, meta-learning has been widely employed to train general models via episodic learning. Early work~\cite{wang2019panet} constructs class prototypes by applying masked average pooling (MAP) to support images, and segments query images by measuring pixel-wise distances to these prototypes in the metric space. To overcome the limitations of single prototype, \cite{liu2020part,yang2020prototype,zhang2021self,gao2022drnet,wang2023pcnet,chen2024dual} represent each class with multiple prototypes, while \cite{zhang2020sg} employs an additional independent encoder to process MAP features. PCNet~\cite{wang2023pcnet} introduces a self-distillation mechanism to enable complementary learning between query and support prototypes, resulting in more accurate prototype representations.  Since prototype-based methods are inherently limited by feature averaging, subsequent approaches~\cite{liu2021harmonic,zhang2021few,peng2023hierarchical,zhuge2021deep,iqbal2022msanet,min2021hypercorrelation,moon2023msi} explicitly build support-query feature aggregation modules to fully exploit the information in limited supports. 
Among them, \cite{min2021hypercorrelation} first calculates 4D correlation volume between the support and query image. For computational efficiency, this volume is decoupled into 2D similarity maps along the query and support dimensions, alternatively, and then processed with convolutions.
Methods such as~\cite{zhang2021few,lu2021simpler,peng2023hierarchical,shi2022dense,xu2023self,chang2024drnet} adopt support-query cross-attention mechanisms. 
CyCTR~\cite{zhang2021few} consists of two transformers: a self-alignment block that aggregates contextual information within the query image, and a cross-alignment block that integrates pixel-level support features into the query features. 
Unlike~\cite{zhang2021few}, which applies self-attention only to query features, \cite{chang2024drnet} applies self-attention to both support and query features and then uses cross-attention to reorganize these features, generating prototypes for guiding segmentation. Based on support-query fusion, \cite{luo2023pfenet++} performs region matching to produce context-aware prior masks. Other works~\cite{peng2023hierarchical,shi2022dense,zhuge2021deep,iqbal2022msanet,hong2022cost} employ multi-layer features for matching. 
Among them, \cite{peng2023hierarchical} applies knowledge distillation to enable lower-level correlation volumes learned from higher-level maps, capturing richer contextual information. PCN~\cite{lu2023prediction}, on the other hand, proposed normalized score fusion to eliminate directional bias between base and novel classes. It further introduces a transformer based on feature-score cross-covariance for prediction calibration, thereby alleviating the overconfidence issue in novel class predictions. This work provides valuable insights into prediction alignment.
Unlike these approaches, for support-query aggregation, we introduce a Mamba-based module enhanced by contrastive hypercorrelation. 
Building on Mamba’s computational efficiency and its dynamic, input-selective contextual processing capabilities, our method is shown to achieve more comprehensive and robust hypercorrelations for mask decoding. 

\subsection{Vision Foundation Models in Few-shot Segmentation}
With the development of pre-trained visual models such as ViT~\cite{dosovitskiy2020image}, CLIP~\cite{radford2021learning}, SAM~\cite{kirillov2023segment}, and DINOv2~\cite{oquab2024dinov2}, 
Few-shot Semantic Segmentation (FSS) models~\cite{ liu2024matcher,chen2024visual,sun2024vrp}  have achieved significant improvements on benchmark datasets. 
Early work, CWT~\cite{lu2021simpler}, demonstrated that large-scale pre-training enables fine-tuning-free transfer to few-shot segmentation, with the classifier adaptively learned via a classifier weight transformer. Its underlying idea is closely aligned with the current paradigm of leveraging vision foundation models for few-shot segmentation.
Later work~\cite{wang2024rethinking} casts a CLIP on top of existing well-trained FSS networks, leading to further improvements. The training-free methods~\cite{zhang2023personalize, liu2024matcher} pioneer a SAM-based framework for FSS, where Matcher~\cite{liu2024matcher} generates point prompts using support-query similarity based on DINOv2 features, with dual-directional pruning. Inspired by~\cite{liu2024matcher},~\cite{zhang2025bridge} enhances segmentation by building a graph on the generated masks. Other works, like~\cite{sun2024vrp}, aim to learn prompt encoders to avoid complex hand-crafted procedures but still rely on both SAM and DINOv2, resulting in large memory footprints. 
SegIC\cite{meng2024segic} bypasses SAM and trains a decoder directly on DINOv2 features, while also introducing a CLIP text encoder to generate meta-prompts. 
SAM2\cite{ravi2025sam} further improves upon SAM by inheriting its general-purpose segmentation capability while enhancing temporal modeling and video segmentation performance, leading to better overall accuracy and generalization. 
SANSA~\cite{cuttano2026sansa} observes that SAM2's representations are entangled with object-tracking cues, impairing semantic understanding. It adopts the SAM2 full model and introduces lightweight adapters ($\sim$10M params) to disentangle semantics and shift Memory Attention from visual to semantic similarity, thereby effectively adapting SAM2 to few-shot segmentation tasks. 
FSSAM~\cite{xu2025unlocking} adopts a dual-encoder architecture, using SAM2 for segmentation and DINOv2 for feature extraction. Specifically, it designs a pseudo prompt generator that calculate the 2D distance map between the support prototype and query features on DINOv2 for prompting.  It further incorporates the iterative memory refinement built on SAM2 to progressively integrate query foreground features, along with a support-calibrated memory attention to suppress background responses. 
Compared with these dual-encoder architectures, our method achieves significantly superior performance with higher efficiency.
This primarily stems from the pivotal finding of  feature alignment between DINOv2's low-level features and SAM encoder's outputs, which enables a superior decoder through embedding distillation. 
Recent work RADIOv2.5~\cite{heinrich2025radiov2} performs multi-teacher knowledge agglomeration for general-purpose visual representation learning. It trains a full ViT backbone with separate adapter heads for features from SAM, DINOv2, and CLIP on DataComp1B, focusing on resolution robustness and token compression across diverse tasks. In contrast, our UINO-FSS targets few-shot segmentation via single-teacher distillation from DINOv2 to SAM, leveraging early feature consistency to align with SAM's decoder. It trains only a lightweight adapter (1M params) on 1\% of SA-1B data, achieving high efficiency.

\subsection{Mamba-based Architectures in Vision Tasks}
Mamba~\cite{gu2023mamba} is widely used in vision tasks owing to its linear complexity and strong long-range dependency modeling ability.
HMNet~\cite{xu2024hybrid} for few-shot segmentation is a hybrid Mamba network that uses self Mamba blocks to model intra-image dependencies and hybrid Mamba blocks to periodically re-scan support features for efficient support–query fusion.
To tackle Mamba’s limitations in 2D spatial modeling, LocalMamba~\cite{huang2024localmamba} enhances visual representations via window-based local scanning and adaptive direction search, while Spatial-Mamba~\cite{xiao2025spatial} adopts a structure-aware state fusion equation with dilated convolutions to capture spatial dependencies and reduce context loss from unidirectional scanning.
For dynamic modeling, DefMamba~\cite{liu2025defmamba} strengthens features with a multi-scale backbone and deformable Mamba blocks; GroupMamba~\cite{shaker2025groupmamba} designs a Modulated Group Mamba Layer with channel grouping and affinity modulation to build efficient backbones.
MambaPanoptic~\cite{cheng2026mambapanoptic} extends Mamba to panoptic segmentation using MambaFPN and PanopticFCN for efficient global multi-scale modeling and proposal-free prediction.
VMamba\cite{liu2024vmamba} comprises stacked Visual State-Space (VSS) blocks centered on the 2D Selective Scan (SS2D) module. SS2D uses four-directional scanning to bridge 1D sequential and 2D visual structures for context aggregation.
Unlike existing Mamba-based methods that model raw 2D images, the proposed Mamba-HyperCorrelation Module (MHCM) in UINO-FSS models the support-query 4D hypercorrelation volume with patch-wise Visual State Space (VSS), and alternately employs convolutions and VSS for global hypercorrelation modeling to capture fine-grained correspondences.

\begin{figure*}[htbp]
\centering
\includegraphics[width=1\textwidth]{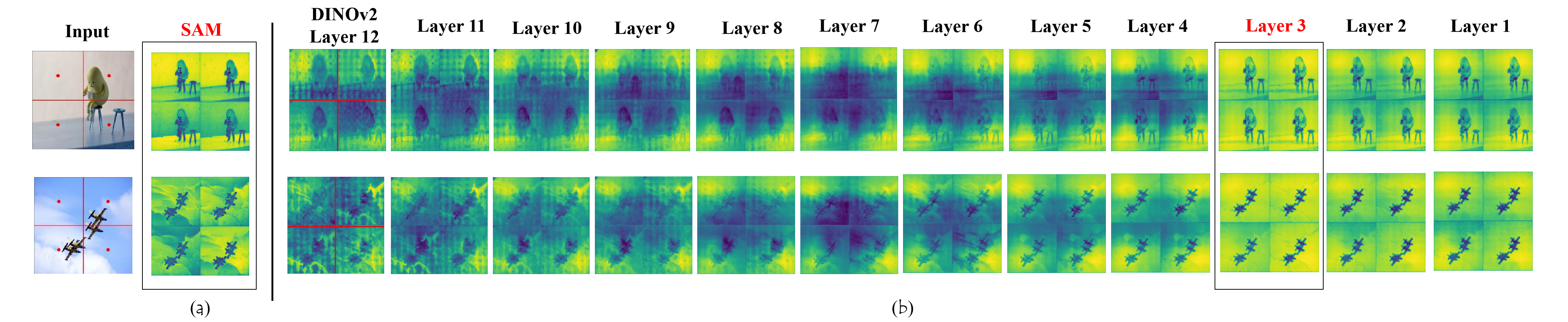}
\caption{\textbf{Analyzation on the embeddings from DINOv2 and SAM}. We compute the self-similarity map (SSM) of feature vectors on red dots to analyze the embedding distribution of both models and visualize them in the figure. Here, (a) shows the SSM of features from the last layer of SAM's encoder, while (b) displays the SSM of features from all layers of DINOv2-Base. Unlike SAM's holistic semantic focus, DINOv2's high-level embeddings concentrate on various local regions, offering richer representation. Despite this, we found that embeddings from DINOv2's 3rd layer are most similar to SAM's encoder embeddings. This observation enables efficient cross-model distillation for our lightweight segmenter.  }
\label{fig:analyze-dino-sam-feat}
\vspace{-7pt}
\end{figure*}

\section{Preliminary}
\label{sec:Preliminary}
While existing methods~\cite{liu2024matcher,zhang2025bridge,sun2024vrp} successfully harness the complementary strengths of SAM and DINOv2 by employing dual encoders for independent feature extraction, this design also leads to models with substantial parameter counts and high inference time. To address this issue, we propose learning SAM's feature distribution by introducing a small number of trainable parameters after DINOv2 encoder, enabling SAM's powerful decoder on top of DINOv2. 
However, experimental results reveal significant distinctions between the features from these two models, as illustrated in Fig.~\ref{fig:analyze-dino-sam-feat}. The distillation loss that aims to align the two kinds of features decreases slowly. To this end, we compute self-similarities of features from DINOv2 and SAM's encoders separately for analysis.

Given a feature representation $F \in \mathbb{R}^{H \times W \times C}$ obtained from the image encoder, we first reshape it into a two-dimensional matrix $F \in \mathbb{R}^{H W \times C}$, and then compute the cosine similarity after normalization, formulated as:
\begin{equation}
S_{ij} = \frac{\langle F_i, F_j \rangle}{\|F_i\| \|F_j\|}, \quad \forall i,j \in [1, HW],
\label{eq:SSM}
\end{equation}
where $S_{ij}$ represents the similarity between feature vectors at positions $i$ and $j$. For each position $i$, by computing its correlation with all positions in the image, we get a similarity map of size $H\times W$. Fig.~\ref{fig:analyze-dino-sam-feat} shows instances of the Self-Similarity Maps (SSM) on four anchor points.

As illustrated in Fig.~\ref{fig:analyze-dino-sam-feat}, the self-similarity analysis reveals distinct characteristics between the representations of SAM and DINOv2. SAM’s features exhibit strong global characteristics, where each pixel can perceive semantic information across the entire image. 
In contrast, DINOv2's features are more localized, with each pixel focusing primarily on the neighboring regions of the same semantics. \textit{An interesting observation} is that DINOv2 can already capture holistic attention in its early-stage embeddings (Layer 3 in Fig.~\ref{fig:analyze-dino-sam-feat} (b)) but is trained to represent richer details in the high-level layers. \textbf{This suggests that a complete DINOv2 model is not strictly necessary for capturing SAM's representation, and transferring knowledge from DINOv2's early-stage features is practicable.} For DINOv2-Base, we utilize features from the 3rd layer, which are found to properly match the distribution of SAM's output embedding.

\begin{figure}[t]
    \centering
    \includegraphics[width=1.0\linewidth]{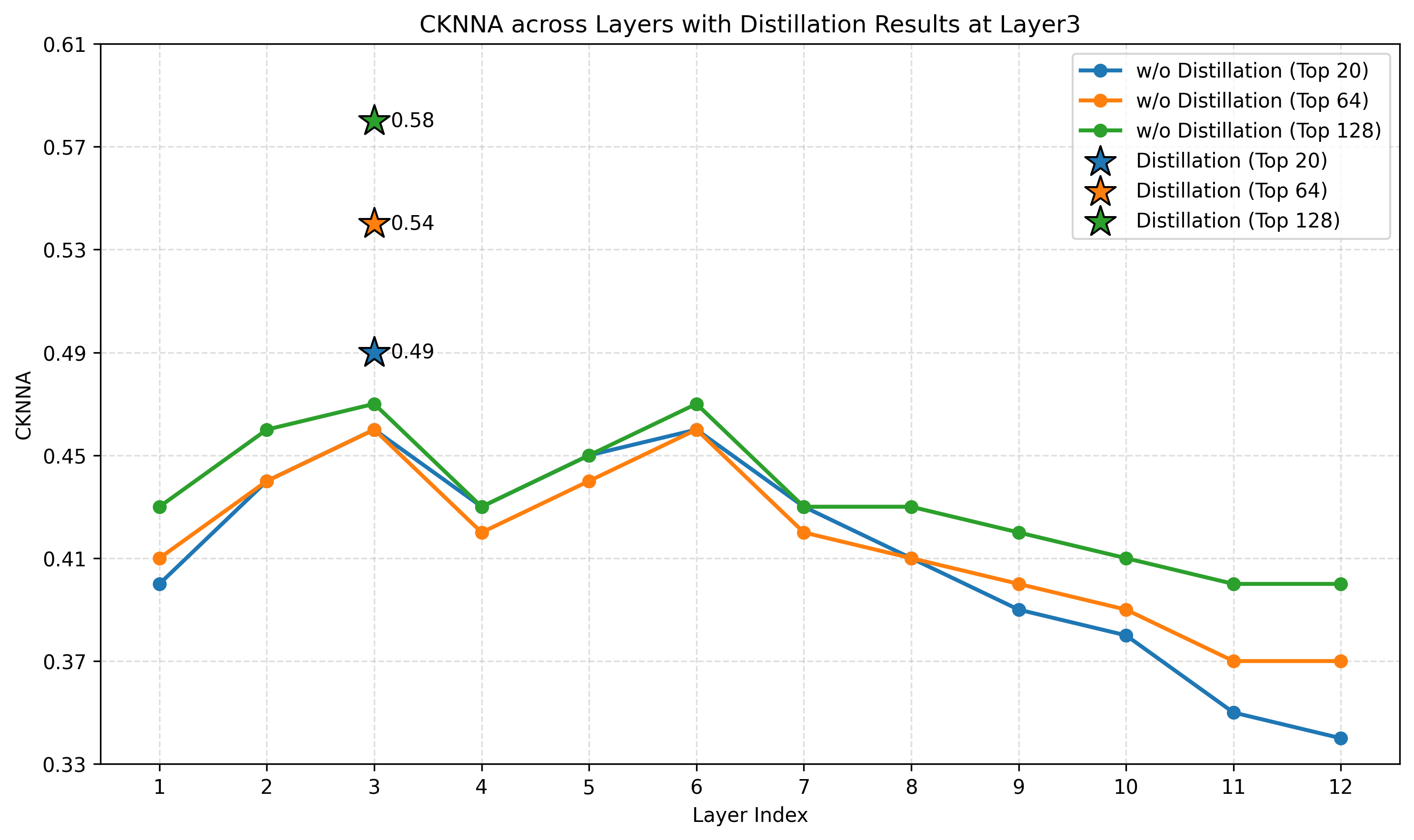}
    \caption{\textbf{CKNNA values between features from various DINOv2 layers and the SAM Encoder.} The star markers denote the CKNNA results after distillation. Specifically, \textit{w/o Distillation} indicates the CKNNA obtained on original DINOv2 embeddings, while \textit{Distillation} presents the scores on DINOv2 embeddings learned with distillation.} 
    \label{fig:cknna_results}
\end{figure}

Motivated by the Centered Kernel Nearest-Neighbor Alignment (CKNNA)~\cite{cknna}, which measures whether two representation spaces exhibit similar local similarity structures, we extend SSM further to a scalable metric. Specifically, we compute the CKNNA between the features of DINOv2 Layers 1-12 and SAM Encoder's output, to further justify the selection of the DINOv2's third layer for distillation.

To adapt CKNNA for measuring the distribution consistency of local features, we first compute the self-similarity matrices for all patches in the image, yielding the full SSM $\mathbf{S}^k_{dinov2} \in \mathbb{R}^{HW\times HW}$ on the $k$-th layer of DINOv2 and $\mathbf{S}_{sam} \in \mathbb{R}^{HW\times HW}$ on SAM's encoder output, where H and W denote the height and width of the feature map. Each row of $\mathbf{S}$ represents the self-similarities of a specific patch to all the others. For each patch in the image, i.e, a position in the feature map, we select the top-$t$ nearest neighbors, obtaining index set $P^k_{dinov2,i}$ and $P_{sam,i}$ for the $i$-th patch. We then compute the intersection set $\hat{P}^k_{i}$. Within this intersection set, we retain the original scores from $\mathbf{S}^k_{dinov2}$ and $\mathbf{S}_{sam}$ 
at the corresponding indices, and set all other entries to zero. After centering the similarity values, we obtain the new similarity maps, denoted as $\mathbf{\hat{S}}^k_{dinov2}$ and $\mathbf{\hat{S}}_{sam}$. CKNNA is defined as:
\begin{equation}
\begin{aligned}
& \mathrm{CKNNA}(\mathbf{\hat{S}}_{dinov2}^k,\mathbf{\hat{S}}_{sam}) = \\
& \frac{\mathrm{Align}(\mathbf{\hat{S}}_{dinov2}^k,\mathbf{\hat{S}}_{sam})}
{\sqrt{\mathrm{Align}(\mathbf{\hat{S}}_{dinov2}^k,\mathbf{\hat{S}}_{dinov2}^k)\cdot \mathrm{Align}(\mathbf{\hat{S}}_{sam},\mathbf{\hat{S}}_{sam})}} ,
\end{aligned}
\end{equation}
where $\mathrm{Align}(\cdot,\cdot)$ denotes a localized Hilbert-Schmidt Independence Criterion (HSIC)~\cite{hsic}, where HSIC measures statistical dependence between variables in a reproducing kernel Hilbert space.

The statistics of resulting CKNNA scores between DINOv2 Layers 1-12 and the SAM encoder's output, gained on 5000 images sampled from SA-1B dataset~\cite{kirillov2023segment},  are shown in Fig.~\ref{fig:cknna_results}. As can be observed, early- and middle-stage DINOv2 features generally show higher local structural consistency with the final SAM feature, while deeper layers exhibit gradually decreasing  CKNNA scores. This suggests that deeper semantic features are not fully aligned with SAM's dense prediction representation. In particular, Layer 3 and Layer 6 achieve relatively high scores under different top-$t$ settings, indicating that they better preserve local spatial structures related to SAM representations.

\vspace{-2pt}
\section{Approach}
\subsection{Problem Definition}
In the task of Few-Shot Semantic Segmentation (FSS), a dataset $D$ is divided into a training set $D_{\text{train}}$ and a test set $D_{\text{test}}$, with disjoint category sets, i.e., $C_{\text{train}} \cap C_{\text{test}} = \emptyset$, to ensure no class overlap between training and testing. The FSS model is first trained on the base classes from the training set and is then directly evaluated on the novel classes in the test set, without any additional training or fine-tuning. Specifically, given $N$ support images and their corresponding pixel-wise masks for a novel class $c$, the FSS model aims to segment the pixels belonging to class $c$ in a query image based on the support information. This process is referred to as N-shot segmentation.

\begin{figure*}[htbp]
\centering
\includegraphics[width=1\textwidth]{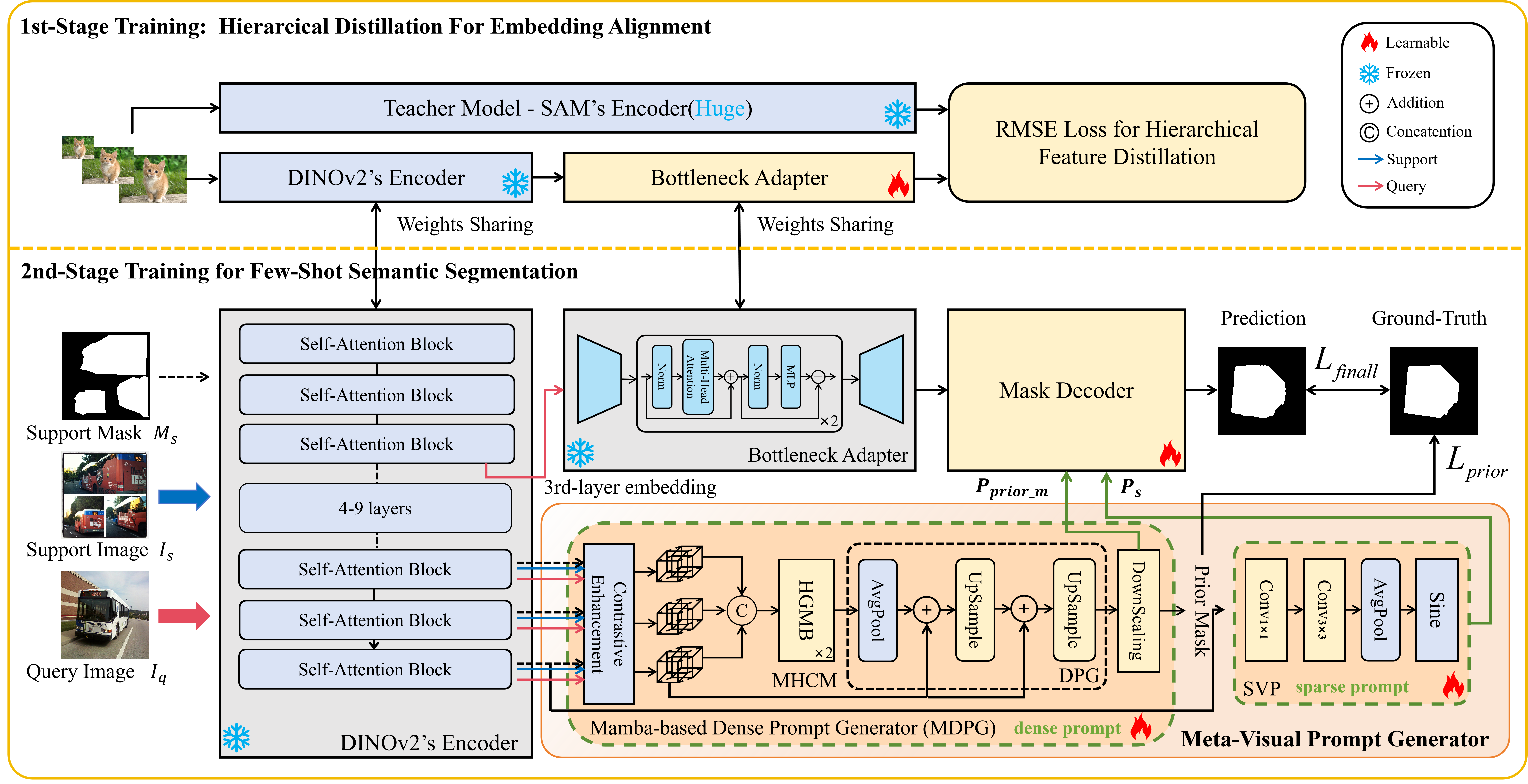}
\caption{\textbf{The proposed UINO-FSS architecture.}  Our architecture consists of a DINOv2 encoder and a lightweight segmenter that includes a bottleneck adapter (BA), a Meta-Visual Prompt Generator (MVPG) and a mask decoder. Upper part is the coarse-to-fine cross-model distillation procedure, with only the adapter trainable for feature matching. Below is the overall architecture of our few-shot semantic segmentation model. Our MVPG includes two modules for the Semantic-aware Visual Prompts (SVP) and the Mamba-based dense prompts, respectively. }
\label{fig:framework}
\vspace{-7pt}
\end{figure*}
\subsection{Model Overview}
In this work, we introduce a unified framework consisting of a DINOv2 encoder and a lightweight segmenter, as shown in Fig.~\ref{fig:framework}. In the following, we will give an overview of both our architecture and the training strategy.

\paragraph{The Architecture} 
The unified UINO-FSS comprises a frozen DINOv2's encoder and a lightweight segmenter, as shown in Fig.~\ref{fig:framework}. Inspired by SAM~\cite{kirillov2023segment}, the segmenter is designed to decode masks using both prompts and image features. Unlike SAM encoding points/boxes, our segmenter directly generates visual prompts. In particular, the segmenter incorporates three components: a Bottleneck Adapter (BA), a Meta-Visual Prompt Generator (MVPG), and a mask decoder. The BA imports the low-level features from the frozen encoder and adapts them into embeddings suited for boundary localization and semantic segmentation. The MVPG generates semantic-aware visual prompts and Mamba-based dense prompts using features from the encoder's last layer.\footnote{The semantic-aware prompt and Mamba-based dense prompt work as sparse and dense prompts, respectively, as used in SAM.} Inheriting SAM's decoder architecture, our mask decoder integrates prompts with image features generated by the bottleneck adapter to produce the final mask.

\paragraph{Training Strategy} Our training consists of two stages, i.e., the embedding distillation and training for mask decoding. During both training stages, the DINOv2 encoder is frozen and only parts of our segmenter are trained.
\textbf{In the first stage, we train only the adapter with a Root Mean Square Error (RMSE) loss to align the embeddings from SAM and the 3rd layer of DINOv2, without being aware of the downstream task.  
In the second stage, with the DINOv2's encoder and the adapter frozen, the remaining components of the segmenter, i.e., the MVPG and mask decoder, are further optimized via meta learning for few-shot semantic segmentation, where the decoder is initialized with SAM's parameters.}
By integrating knowledge distillation with an effective decoder initialization, our method successfully combines the strengths of DINOv2 and SAM.

\subsection{Hierarchical Distillation for Embedding Alignment} 
\label{sec:distillation}
Based on our observation in Sec.~\ref{sec:Preliminary}, we directly leverage the 3rd-layer features of DINOv2-Base and train a lightweight adapter with a bottleneck structure to align its embeddings with SAM. 
During embedding distillation, SAM-huge's encoder serves as the teacher, while DINOv2's 1-3 layers together with the trainable adapter,  constitute the student. Only the adapter's parameters are updated during the process. The features produced by the learned adapter will be used as the image embedding input to the mask decoder for few-shot segmentation.

\subsubsection{The Bottleneck Adapter}
To maintain a lightweight design, we build a bottleneck-structured adapter to efficiently transfer embedding features. As shown in Fig.~\ref{fig:framework}, the adapter takes features from DINOv2-Base's low-level layer $l_{low}$ ($l_{low}$=3) as input, denoted as $F^{low} \in \mathbb{R}^{H \times W \times C}$, and transforms them into \( F^{ba} \in \mathbb{R}^{H \times W \times C_{sam}} \) to align with the feature dimension of SAM encoder's last layer, where $C=768$,  $C_{sam}=256$, H and W are the height and width of the feature map, respectively\footnote{Given input image size $518\times 518$, we get feature dimensions H and W both being 37.}.
Specifically, we first apply a sequence of convolutions to reduce the channel dimension from \( C \) to a lower intermediate dimension $C'$ ($C'=128$), in order to compress redundant information and reduce computation\footnote{Convolutions with sizes of $1\times1\times768\times256$, $3\times3\times256\times256$, and $1\times1\times256\times128$ are applied here for dimension reduction.}. We then employ two self-attention blocks to enhance the feature representation, followed by a pointwise convolution that increases the channel dimension to the target dimension \( C_{sam} \). The self-attention blocks adopt the standard Transformer structure, with $Q,K,V \in \mathbb{R}^{C'\times C'}$ and the inner-layer dimension of feed-forward network (FFN) being 512, enabling long-range dependency modeling across channels.  For other models within the DINOv2 family, i.e., DINOv2-Small/Large, we employ a similar design to align their low-level features with the output of the SAM encoder\footnote{For DINOv2-Small/Large, we choose comparable low-level layers, i.e., $l_{low} =2$ and 4, respectively. }.

\subsubsection{Cross-model Distillation from Coarse to Fine}
Given that SAM's features capture holistic semantic information, we design a hierarchical distillation procedure to progressively align features from coarse to fine across multiple stages. Specifically, we start training with a low input resolution of $126\times 126$, and at each distillation stage, the resolution is doubled for further training. This iterative process is conducted twice, eventually reaching a resolution of $518\times 518$, which corresponds to the input size used by DINOv2. Through this progressive training strategy, our model gradually fits the feature distribution of SAM and effectively replaces its image encoder. In practice, due to SAM's holistic-embedding nature, a coarse-to-fine distillation over hierarchical resolutions proves critical, as direct distillation on fine resolutions leads to slow convergence.
Let's denote the output of the bottleneck adapter as \( F^{ba} \), and the final-layer features from the encoder of SAM-ViT-Huge as \( F^{sam} \). By minimizing the Root Mean Square Error (RMSE) loss between \( F^{ba} \) and \( F^{sam} \), we encourage our model to match the output distribution of SAM. Specifically, the distillation objective minimizes the $\ell_2$ distance between features of corresponding patches across the two encoders (DINOv2 and SAM). This iterative alignment of local feature distributions ultimately promotes global consistency. Fig.\ref{fig:cknna_results} shows that CKNNA scores on embeddings from the distilled adapter are substantially higher than those from the original embeddings without distillation, further validating the rationale of the proposed hierarchical distillation. Notably, only 1\% of SAM's training images from SA-1B dataset, are used throughout the distillation process.

\subsection{Meta-Visual Prompt Generator}

Following the architecture of SAM's mask decoder, our decoder for few-shot segmentation takes three inputs: the image embedding, the dense prompt, and the sparse prompt. Specifically, the output features from the bottleneck adapter (Section~\ref{sec:distillation}) serve as the image embedding. The proposed Meta-Visual Prompt Generator (MVPG) provides both the dense prompt via a Mamba-based dense prompt generator, and the sparse prompts, which are realized as semantic-aware visual prompts. Notably, instead of using early-stage features of DINOv2 as the image embedding, the MVPG operates on support and query features extracted from the last layer of DINOv2 for matching and prompting. This design choice is motivated by two complementary considerations on DINOv2: its early features exhibit better global consistency with the output of SAM's encoder, while deeper features offer richer fine-grained correspondences, making them more suitable for matching and discrimination. 

\subsubsection{Semantic-aware Visual Prompts}
To provide Semantic-aware Visual Prompts (SVP), we directly generate embedding vectors from the support features. Specifically, we first perform dimension reduction on the last layer feature map $F \in \mathbb{R}^{H_s \times W_s \times C}$ of DINOv2 through convolutional layers to obtain $F^{'} \in \mathbb{R}^{H_s \times W_s \times C_{sam}}$,
\begin{equation}
F^{'} = \mathcal{F}_{\text{conv}}(F),
\end{equation}
where $\mathcal{F}_{\text{conv}}$ denotes a convolutional block composed of a $1 \times 1$ and $3 \times 3$ convolution, each followed by layer normalization. We then perform masked average pooling (MAP) on $F^{'}$ to extract the foreground class prototypes. 
To align with SAM's positional encoding, we further encode these prototypes using the sine function to obtain the sparse prompts $P_s \in \mathbb{R}^{1 \times C_{sam}}$,
\begin{equation}
P_s = \mathcal{F}_{\text{sine}}(\mathcal{F}_{\text{MAP}}(F^{'})).
\end{equation}
The semantic-aware visual prompt $P_s$ matches the dimension of SAM's sparse prompt for direct use with its decoder.

\subsubsection{Mamba-based Dense Prompt Generator}
By integrating the SVP and a vanilla 2D similarity map between query features and support prototypes, we can already construct a strong framework with negligible parameters. To further improve the framework while maintaining its lightweight structure, we introduce the Mamba-based dense prompt generator, featuring the contrastive enhancement on similarity volumes, a Mamba-HyperCorrelation Module (MHCM), and a module for dense prompt generation.

\begin{figure}[t]
\centering
\includegraphics[width=0.35\textwidth]{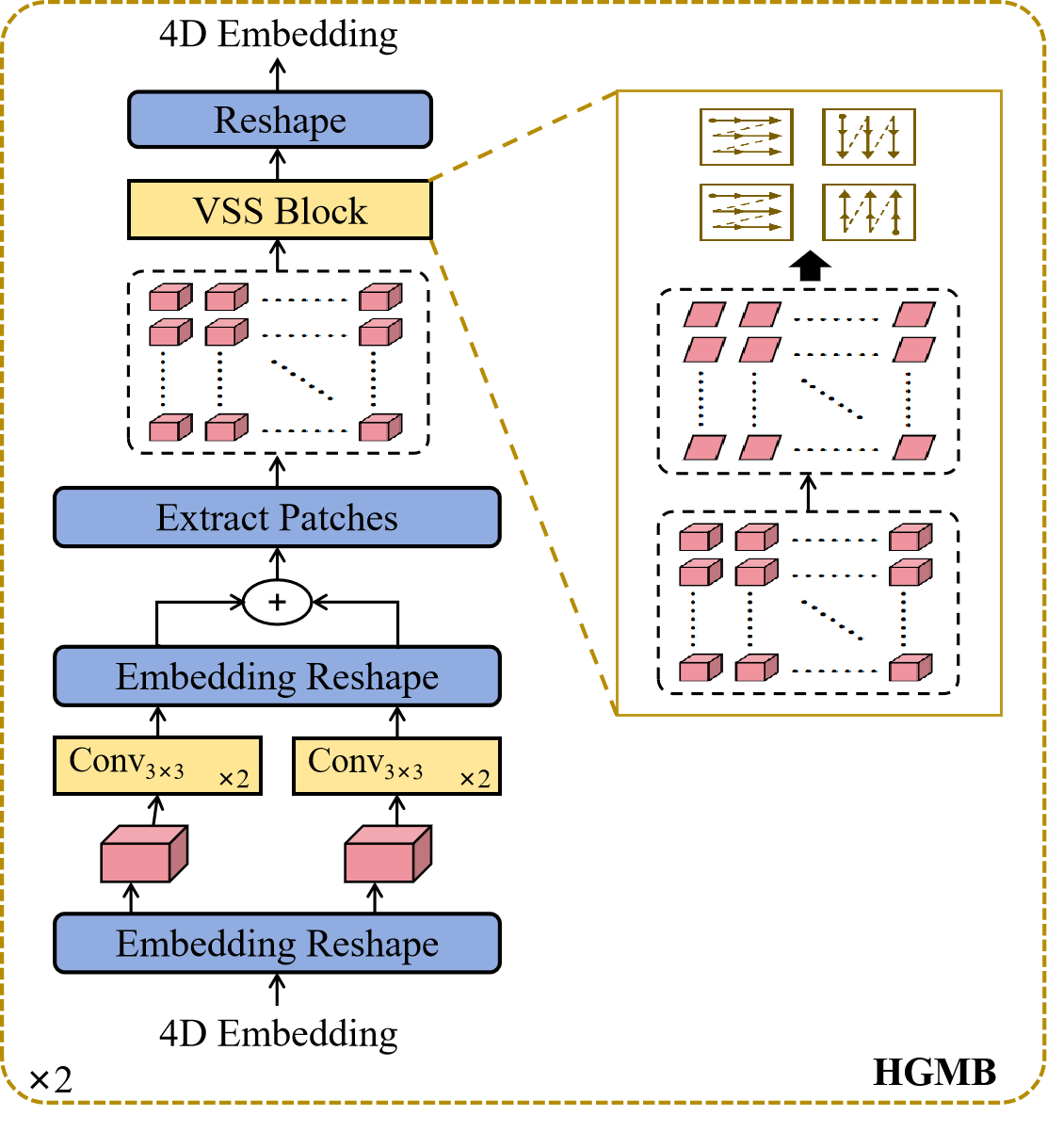}
\caption{\textbf{Network structure for the Mamba-HyperCorrelation Module (MHCM). } MHCM stacks two Hierarchical Global Modeling Blocks (HGMB), which process 4D volumetric correlation while maintaining high efficiency. }
\label{fig:HGMB}
\end{figure}

\paragraph{Contrastive Enhancement} This module is proposed to inject contrastive embeddings into correlation volumes, to enhance subsequent hypercorrelation modeling.  
To this end, we extract features from the last three layers ($l=3$) of the DINOv2 encoder and embed the query's similarities to both the support's foreground and background regions.

Specifically, we first perform an element-wise multiplication between the support mask, $M_s \in \{0,1\}^{H_s \times W_s}$, and the support features, achieving the masked foreground features $F_s^{fg}$ and background features $F_s^{bg}$  as below, 
\begin{equation}
F_s^{fg} = M_s \odot F_s, \quad F_s^{bg} = (1 - M_s) \odot F_s.
\end{equation}
For each selected layer, we compute two correlation volumes between the query features and the masked support features:
\begin{equation}
\mathcal{V}_{fg}^{(i)}{}_{h_q, w_q, h_s, w_s} = 
\frac{\langle F_q^{(i)}(h_q, w_q), F_s^{fg(i)}(h_s, w_s) \rangle}{\| F_q^{(i)}(h_q, w_q)\| \cdot \|F_s^{fg(i)}(h_s, w_s)\|},
\label{eq:vfg}
\end{equation}
\begin{equation}
\mathcal{V}_{bg}^{(i)}{}_{h_q, w_q, h_s, w_s} = 
\frac{\langle F_q^{(i)}(h_q, w_q), F_s^{bg(i)}(h_s, w_s) \rangle}
{\|F_q^{(i)}(h_q, w_q)\| \cdot \|F_s^{bg(i)}(h_s, w_s)\|}.
\end{equation}
where $\mathcal{V}_{fg}^{(i)}$ and $\mathcal{V}_{bg}^{(i)} \in \mathbb{R}^{H_q \times W_q \times H_s \times W_s}$ denote the correlation volumes derived from the $i$-th layer, computed using the foreground and background support features, after which ReLU is applied to both volumes.
To suppress the inherent background noise of the ViT backbone, we directly subtract the background correlation volume from the foreground correlation volume on each of the last $l$ layers and concatenate the results. This process can be formulated as below,
\begin{equation}
\mathcal{V}^{(i)} = \mathcal{V}_{fg}^{(i)} - \mathcal{V}_{bg}^{(i)}.
\end{equation}
\begin{equation}
\mathcal{V} = \mathcal{V}^{(L-l+1)} \oplus \mathcal{V}^{(L-l+2)} \oplus \dots \oplus \mathcal{V}^{(L)},
\end{equation}
where $L$ denotes the overall number of layers in DINOv2 image encoder, the symbol $\oplus$ indicates concatenation along an additional dimension. $\mathcal{V} \in \mathbb{R}^{l \times H_q \times W_q \times H_s \times W_s}$ encodes the query-to-support matching information across multiple layers.

\paragraph{Mamba-HyperCorrelation Module} 
To avoid the high cost of 4D convolution, prior work~\cite{min2021hypercorrelation} reshapes the volume $\mathcal{V}$ into forms $\mathcal{V}_{a} \in \mathbb{R}^{(H_s \cdot W_s) \times l \times H_q \times W_q}$ and $\mathcal{V}_{b} \in \mathbb{R}^{( H_q \cdot W_q) \times l \times H_s \times W_s}$, which can then be processed with 2D convolutions independently in batch\footnote{Here, the first dimension always serves as batch dimension. 
}. This strategy, however, lacks global semantic modeling, limiting its ability to correct regional misalignments. To address this, we introduce the Hierarchical Global Modeling Block (HGMB) that processes true 4D volumes while maintaining high efficiency. By stacking HGMBs, we construct the Mamba-HyperCorrelation Module.

Specifically, the HGMB integrates VMamba~\cite{liu2024vmamba}, alternating between local and global modeling by stacking a convolution block and a patch-wise visual state space (VSS) block. Its structure is shown in Fig.~\ref{fig:HGMB}. In detail, (1) we first apply two stacked $3 \times 3$ convolutions to locally process the 2D correlations $\mathcal{V}^{a}$ and $\mathcal{V}^{b}$ to increase the channel dimensionality\cite{min2021hypercorrelation}. 
The results are then reshaped back to the shape of $\mathcal{V}$ and summed together to form a new correlation volume, 
\(\hat{\mathcal{V}} \in \mathbb{R}^{l' \times H_q \times W_q \times H_s \times W_s}\), which integrates interactions within the support and query neighbors; (2) Given the augmented correlation volume $\hat{\mathcal{V}}$ , we partition it into non-overlapping local blocks within the 4D space. This operation reshapes the volume $\hat{\mathcal{V}}$ into $\mathbb{R}^{l' \times N \times k_q \times k_q \times k_s \times k_s}$, where $k_q$ and $k_s$ are partition sizes along query and support dimensions, and N is the number of local blocks\footnote{$N = \frac{H_q}{k_q} \times \frac{W_q}{k_q} \times \frac{H_s}{k_s} \times \frac{W_s}{k_s}$.}. Subsequently, we flatten the query and support spatial dimensions of each block, transforming it into a 2D patch. This yields the representation $\mathcal{V}_p \in \mathbb{R}^{l^' \times N \times k_q k_q \times k_s k_s}$.  To suppress the potential interference caused by noisy or abnormal correlations within these local blocks, we incorporate a Visual State Space (VSS) module to process each 2D patch. Finally, the processed correlation volumes are reshaped back to their original 4D structure, yielding $\widetilde{\mathcal{V}}'$ formulated as below,
\begin{equation}
\widetilde{\mathcal{V}}' = \mathrm{Reshape}(\mathcal{F}_{\text{vss}}(\mathcal{V}_p)).
\label{eq:vss}
\end{equation}
In Eq.~\ref{eq:vss}, the VSS module, denoted by $\mathcal{F}_{\text{vss}}$, processes each 2D patch by traversing and modeling its structure through a structured state space model (SSM). This formulation enables efficient capture of long-range dependencies within the patch while suppressing noise interference, all with linear computational complexity relative to the sequence length.

\paragraph{Dense Prompt Generation}
To generate the dense prompt, we project the refined correlation volume to the decoder's input dimension. 
Specifically, the refined $\widetilde{\mathcal{V}}'$ is first averaged along the support dimensions into a 2D representation, then combined via a learnable weighted sum with a 2D correlation calculated between the support prototype and query features from DINOv2.  
Next, the resultant aggregated correlation map, denoted as \( \mathcal{V}_{2d} \), is upsampled and further projected into a prior mask \( P_{\text{prior\_m}} \in \mathbb{R}^{1\times H'_q \times W'_q} \), where $H'_q$ and $W'_q$ denote the final mask size.
This prior mask, 
when input to the prompt encoder, 
can provide a dense prompt to guide the mask decoder. The process is as below:
\begin{equation}
P_{\text{prior\_m}} = \mathcal{F}_{\text{Proj}}(\mathrm{Upsample}(\mathcal{V}_{2d})),
\end{equation}
where \(\mathrm{Upsample}(\cdot)\) denotes the upsampling operation and $\mathcal{F}_{\text{Proj}}(\cdot)$ is a convolution layer.

\subsection{Mask Decoding via Meta-Visual Prompting}
Given query image features from bottleneck adapter and prompts from the meta-visual prompt generator, the decoder predicts masks corresponding to the support class. To this end, the mask decoder is first initialized via parameters from the pre-trained mask decoder of SAM-ViT-Huge, and then trained via an episodic training strategy. This training strategy is widely used in meta-learning to simulate the inference process of few-shot segmentation, enabling effective generalization to unseen classes. Specifically, we construct each image batch via several support-query pairs following~\cite{peng2023hierarchical} and learn to predict masks for the current support class.
It is important to note that during episodic training, we train exclusively on the base classes without involving the novel classes. 
The Dice Loss and BCE Loss are computed separately for the prior mask \( P_{\text{prior\_m}} \) and the final prediction \( P_m \)  against the ground truth mask \( M_q \). Particularly, the loss functions are defined as:  
\begin{equation}  
\mathcal{L}_{\text{prior}} = \mathcal{L}_{\text{Dice}}(P_{\text{prior\_m}}, M_q) + \mathcal{L}_{\text{BCE}}(P_{\text{prior\_m}}, M_q),
\end{equation}  
\begin{equation}  
\mathcal{L}_{\text{final}} = \mathcal{L}_{\text{Dice}}(P_m, M_q) + \mathcal{L}_{\text{BCE}}(P_m, M_q),
\end{equation}  
\begin{equation}  
\mathcal{L} = \mathcal{L}_{\text{prior}} + \mathcal{L}_{\text{final}},
\end{equation}  
where \( \mathcal{L}_{\text{Dice}}(\cdot) \) and \( \mathcal{L}_{\text{BCE}}(\cdot) \) represent the Dice Loss and Binary Cross-Entropy (BCE) Loss, respectively.

\section{Experiments}
\subsection{Datasets and Evaluation Metrics}
To evaluate the effectiveness of our proposed method, we conducted extensive experiments on two widely recognized benchmark datasets within the Few-Shot Segmentation (FSS) setting: PASCAL-5$^i$ \cite{shaban2017one} and COCO-20$^i$ \cite{nguyen2019feature}. The PASCAL-5$^i$ dataset, derived from PASCAL VOC 2012 \cite{everingham2010pascal}, is augmented with additional annotations from SDS \cite{hariharan2011semantic} and consists of 20 object classes. The COCO-20$^i$ dataset, constructed from MSCOCO \cite{lin2014microsoft}, presents a more challenging setting as it includes 80 classes. For task partitioning, both PASCAL-5$^i$ and COCO-20$^i$ follow the standard cross-validation protocol, where all categories are divided into several non-overlapping folds. Each fold is used in turn as the test set, while the remaining categories serve as the training set, ensuring a fair evaluation of the model’s generalization ability across different categories.  All experiments are conducted under both the 1-shot and 5-shot settings to investigate the model’s performance using varying amounts of support images. In addition, no further finetuning on the novel set is conducted. Furthermore, we conduct an out-of-distribution evaluation on FSS-1000~\cite{li2020fss}, a dataset specifically designed for FSS containing 1k classes. For evaluation, we adopt the standard mean Intersection over Union (mIoU) across various classes, i.e.,  $\text{mIoU} = \frac{1}{n} \sum_{i=1}^{n} \text{IoU}_i$, as the metric, where $n$ is the number of test cases.

\begin{table*}[t]
    \centering
    \small
    \caption{Performance comparisons on PASCAL-5$^i$ under 1-shot and 5-shot settings. Results in bold indicate the best performance, and results with underlining represent the second-best performance.}
    \setlength{\tabcolsep}{3pt}
    \renewcommand{\arraystretch}{1.2}

    \resizebox{\textwidth}{!}{
    \begin{tabular}{l l c c c c c c c | c c c c c}
        \hline
        \multirow{2}{*}{Method} & \multirow{2}{*}{Backbone} 
        & \multirow{2}{*}{\makecell{Trainable\\Params(M)}} 
        & \multirow{2}{*}{\makecell{Total\\Params(M)}} 
        & \multicolumn{5}{c|}{1-shot} 
        & \multicolumn{5}{c}{5-shot} \\
        \cline{5-14}
        & & & & Fold0 & Fold1 & Fold2 & Fold3 & Mean 
        & Fold0 & Fold1 & Fold2 & Fold3 & Mean \\
        \hline

        BAM $_{CVPR'22}$ \cite{lang2022learning} & \multirow{3}{*}{ResNet50} & 4.9 & 51.7 
        & 69.0 & 73.6 & 67.6 & 61.1 & 67.8 
        & 70.6 & 75.1 & 70.8 & 67.2 & 70.9 \\

        HDMNet $_{CVPR'23}$ \cite{peng2023hierarchical} &  & 4.2 & 50.9 
        & 71.0 & 75.4 & 68.9 & 62.1 & 69.4 
        & 71.3 & 76.2 & 71.3 & 68.5 & 71.8 \\

        AENet $_{ECCV’24}$ \cite{xu2024eliminating} &  & 4.4 & 51.2 
        & 71.3 & 75.9 & 68.6 & 65.4 & 70.3 
        & 73.9 & 77.8 & 73.3 & 72.0 & 74.2 \\ 

        OCNet $_{ICCV'25}$ \cite{wen2025object} &  & - & - 
        & 73.5 & 75.9 & 71.1 & 64.9 & 71.4 
        & 75.9 & 77.1 & 74.1 & 70.9 & 74.5 \\

        \hline

        MSI $_{ICCV’23}$ \cite{moon2023msi} & \multirow{3}{*}{ResNet101} & 3.2 & 47.8 
        & 73.1 & 73.9 & 64.7 & 68.8 & 70.1 
        & 73.6 & 76.1 & 68.0 & 71.3 & 72.2 \\

        SCCAN $_{ICCV’23}$ \cite{xu2023self} &  & 11.4 & 53.9 
        & 70.9 & 73.9 & 66.8 & 61.7 & 68.3 
        & 73.1 & 76.4 & 70.3 & 66.1 & 71.5 \\

        ABCB $_{CVPR’24}$ \cite{zhu2024addressing} &  & - & - 
        & 73.0 & 76.0 & 69.7 & 69.2 & 72.0 
        & 74.8 & 78.5 & 73.6 & 72.6 & 74.9 \\

        \hline\hline

        Matcher $_{ICLR’24}$ \cite{liu2024matcher} & \multirow{2}{*}{\makecell{DINOv2-L, SAM-H}} & 0 & 945.5 
        & 67.7 & 70.7 & 66.9 & 67.0 & 68.1 
        & 71.4 & 77.5 & 74.1 & 72.8 & 74.0 \\

        GF-SAM $_{NeurIPS’24}$ \cite{zhang2025bridge} &  & 0 & 945.5 
        & 71.1 & 75.7 & 69.2 & \underline{}{73.3} & 72.1 
        & \underline{}{81.5} & \underline{}{86.3} & \underline{79.7} & \underline{}{82.9} & \underline{}{82.6} \\

        \hline

        VRP-SAM $_{CVPR’24}$ \cite{sun2024vrp} & \multirow{2}{*}{\makecell{ResNet50, SAM-H}} & 1.6 & 666.4 
        & 73.9 & 78.3 & 70.6 & 65.0 & 71.9 
        & - & - & - & - & - \\

        FCP $_{AAAI’25}$ \cite{park2025foreground} &  & 2.9 & 667.6 
        & 74.9 & 77.4 & 71.8 & 68.8 & 73.2 
        & 77.2 & 78.8 & 72.2 & 67.7 & 74.0 \\

        \hline

        PI-CLIP $_{CVPR’24}$ \cite{wang2024rethinking} & ResNet50, CLIP & 153.8 & 200.5 
        & \underline{}{76.4} & \underline{83.5} & \underline{}{74.7} & \underline{}{72.8} & \underline{}{76.8} 
        & 76.7 & 83.8 & 75.2 & 73.2 & 77.2 \\

        \hline

        FSSAM $_{ICML’25}$ \cite{xu2025unlocking} & DINOv2-B,SAM2-S & 11.5 & 132.6 
        & \textbf{81.6} & \textbf{84.9} & \textbf{81.6} & \underline{76.0} & \textbf{81.0} 
        & \underline{84.1} & \textbf{88.5} & \textbf{83.8} & \underline{85.0} & \textbf{85.4} \\

        \hline

        UINO-FSS (Ours) & DINOv2-B & 4.6 & 92.5 
        & \underline{80.7} & \underline{}{83.2} & \underline{77.8} & \textbf{80.5} & \underline{80.6} 
        & \textbf{84.4} & \underline{87.3} & \underline{}{79.0} & \textbf{85.5} & \underline{84.1} \\
        \hline
    \end{tabular}
    }

    \label{tab:pascal}
\end{table*}

\begin{table*}[t]
    \centering
    \small
    \caption{Performance comparisons on COCO-20$^i$ under 1-shot and 5-shot settings. Our method consistently achieves the best performance, outperforming other approaches based on DINOv2-Base (DINOv2-B) as well as heavier architectures combining DINOv2-Large (DINOv2-L) and SAM-ViT-H (SAM-H).}
    \setlength{\tabcolsep}{3pt}
    \renewcommand{\arraystretch}{1.2}

    \resizebox{\textwidth}{!}{
    \begin{tabular}{l l c c c c c c c | c c c c c}
        \hline
        \multirow{2}{*}{Method} & \multirow{2}{*}{Backbone} 
        & \multirow{2}{*}{\makecell{Trainable\\Params(M)}} 
        & \multirow{2}{*}{\makecell{{Total}\\Params(M)}} 
        & \multicolumn{5}{c|}{1-shot} 
        & \multicolumn{5}{c}{5-shot} \\
        \cline{5-14}
        & & & & Fold0 & Fold1 & Fold2 & Fold3 & Mean 
        & Fold0 & Fold1 & Fold2 & Fold3 & Mean \\
        \hline

        BAM $_{CVPR'22}$ \cite{lang2022learning} & \multirow{3}{*}{ResNet50} & 4.9 & 51.7 
        & 43.4 & 50.6 & 47.5 & 43.4 & 46.2 
        & 49.3 & 54.2 & 51.6 & 49.6 & 51.2 \\

        HDMNet $_{CVPR'23}$ \cite{peng2023hierarchical} &  & 4.2 & 50.9 
        & 43.8 & 55.3 & 51.6 & 49.4 & 50.0 
        & 50.6 & 61.6 & 55.7 & 56.0 & 56.0 \\

        AENet $_{ECCV’24}$ \cite{xu2024eliminating} &  & 4.4 & 51.2 
        & 45.4 & 57.1 & 52.6 & 50.0 & 51.3 
        & 52.7 & 62.6 & 56.8 & 56.1 & 57.1 \\ 

        OCNet $_{ICCV'25}$ \cite{wen2025object} &  & - & - 
        & 45.9 & 56.9 & 52.9 & 50.4 & 51.5 
        & 52.7 & 63.1 & 57.4 & 54.8 & 57.0 \\

        \hline

        MSI $_{ICCV’23}$ \cite{moon2023msi} & \multirow{3}{*}{ResNet101} & 3.2 & 47.8 
        & 44.8 & 54.2 & 52.3 & 48.0 & 49.8 
        & 49.3 & 58.0 & 56.1 & 52.7 & 54.0 \\

        SCCAN $_{ICCV’23}$ \cite{xu2023self} &  & 11.4 & 53.9 
        & 42.6 & 51.4 & 50.0 & 48.8 & 48.2 
        & 49.4 & 61.7 & 61.9 & 55.0 & 57.0 \\

        ABCB $_{CVPR’24}$ \cite{zhu2024addressing} &  & - & - 
        & 46.0 & 56.3 & 54.3 & 51.3 & 51.5 
        & 51.6 & 63.5 & 62.8 & 57.2 & 58.8 \\

        \hline\hline

        Matcher $_{ICLR’24}$ \cite{liu2024matcher} & \multirow{2}{*}{\makecell{DINOv2-L, SAM-H}} & 0 & 945.5 
        & 52.7 & 53.5 & 52.6 & 52.1 & 52.7 
        & 60.1 & 62.7 & 60.9 & 59.2 & 60.7 \\

        GF-SAM $_{NeurIPS’24}$ \cite{zhang2025bridge} &  & 0 & 945.5 
        & 56.6 & 61.4 & 59.6 & 57.1 & 58.7 
        & \underline{67.1} & \underline{}{69.4} & \underline{66.0} & \underline{}{64.8} & \underline{}{66.8} \\

        \hline

        FCP $_{AAAI’25}$  \cite{park2025foreground} & ResNet50, SAM-H & 2.9 & 667.6
        & 46.4 & 56.4 & 55.3 & 51.8 & 52.5 
        & 52.6 & 63.3 & 59.8 & 56.1 & 58.0 \\

        VRP-SAM $_{CVPR’24}$ \cite{sun2024vrp} & DINOv2-B, SAM-H & 1.6 & 666.4 
        & \underline{}{56.8} & 61.0 & \underline{64.2} & \underline{}{59.7} & \underline{}{60.4} 
        & - & - & - & - & - \\

        \hline

        SEGIC $_{ECCV’24}$ \cite{meng2024segic} & DINOv2-B, CLIP & 5.2 & 214.9 
        & 55.8 & 54.7 & 52.4 & 51.4 & 53.6 
        & - & - & - & - & - \\

        PI-CLIP $_{CVPR’24}$ \cite{wang2024rethinking} & ResNet50, CLIP & 153.8 & 200.5 
        & 49.3 & \underline{65.7} & 55.8 & 56.3 & 56.8 
        & 56.4 & 66.2 & 55.9 & 58.0 & 59.1 \\

        \hline

        SANSA  $_{NeurIPS’25}$ \cite{cuttano2026sansa} & SAM2-L & 9.9 & 234.2 
        & - & - & - & - & 60.2 
        & - & - & - & - & 64.3 \\

        \hline

        \hline

        FSSAM $_{ICML’25}$ \cite{xu2025unlocking} & DINOv2-B,SAM2-S & 11.5 & 132.6 
        & \underline{59.9} & \textbf{}{65.6} & \textbf{}{62.1} & \underline{61.6} & \underline{62.3} 
        & \textbf{68.6} & \textbf{74.0} & \underline{}{64.5} & \textbf{69.9} & \underline{69.3} \\

        \hline

        UINO-FSS (Ours) & DINOv2-B & 4.6 & 92.5 
        & \textbf{62.2} & \textbf{66.0} & \textbf{65.5} & \textbf{64.2} & \textbf{64.5} 
        & \underline{}{66.6} & \underline{73.3} & \textbf{70.9} & \underline{68.1} & \textbf{69.7} \\

        \hline
    \end{tabular}
    }
    \label{tab:coco}
\end{table*}

\begin{table}[t]
    \centering
    \small
    \caption{Performance on FSS-1000 under one-shot setting.}
    \setlength{\tabcolsep}{4pt}
    \renewcommand{\arraystretch}{1.3}
    \begin{tabular*}{\linewidth}{@{\extracolsep{\fill}} l c c c c c c @{}}
        \hline
        Metric & 
        \begin{tabular}[c]{@{}c@{}}Painter\\ \cite{wang2023images}\end{tabular} & 
        \begin{tabular}[c]{@{}c@{}}SegGPT\\ \cite{wang2023seggpt}\end{tabular} & 
        \begin{tabular}[c]{@{}c@{}}PerSAM\\ \cite{zhang2023personalize}\end{tabular} & 
        \begin{tabular}[c]{@{}c@{}}PerSAM-F\\ \cite{zhang2023personalize}\end{tabular} & 
        \begin{tabular}[c]{@{}c@{}}SEGIC\\ \cite{meng2024segic}\end{tabular} & 
        Ours \\
        \hline
        mIoU & 61.7 & 85.6 & 71.2 & 75.6 & \underline{86.8} & \textbf{89.0} \\
        \hline
    \end{tabular*}
    \label{tab:zero-shot-rotated}
\end{table}

\begin{table}[t]
    \centering
    \small
    \caption{Additional Complexity ablation study on variants of UINO-FSS and prior methods on COCO-20$^i$ under 1-shot setting. UINO-FSS-Comp(*) refers to our compact model without decoder training. 
    }
    \setlength{\tabcolsep}{4pt}
    \renewcommand{\arraystretch}{1.3}
    \begin{tabular*}{\linewidth}{@{\extracolsep{\fill}} l c c c c @{}}
        \hline
        Method & Train Params & Total Params & mIoU\\
        \hline
        Matcher & 0 & 945.5M & 52.7\\
        GF-SAM & 0 & 945.5M & 58.7\\
        \hline
        \hline
        FCP  & 2.9M & 667.6M & 52.5\\
        SEGIC & 5.2M & \underline{}{214.9M} & 53.6\\
        VRP-SAM (ResNet50)  & \underline{1.6M} & 666.4M & \underline{}{53.9}\\
        \hline
        UINO-FSS-Comp*  & \textbf{0.07M} & \textbf{92.0M} & \underline{}{57.2}\\
        UINO-FSS-Comp  & 4.1M & \textbf{92.0M} & \underline{59.6}\\
        UINO-FSS  & 4.6M & \underline{92.5M} & \textbf{64.5}\\
        \hline
    \end{tabular*}
    \label{tab:ablation-complex}
\end{table}

\begin{table}[h]
    \centering
    \footnotesize
    \caption{Comparison of inference efficiency across methods, including inference time and peak GPU memory usage during testing.}
    \vspace{2pt}
    \setlength{\tabcolsep}{3pt} 
    \renewcommand{\arraystretch}{1.3} 
    \begin{tabular}{c c c c}
        \hline
        Method & Latency (ms) & Peak Memory (M) & mIoU \\
        \hline
        Matcher\cite{liu2024matcher} & 6218.0 & 6936.4 & 52.7 \\
        GF-SAM\cite{zhang2025bridge} & 726.4 & 6910.2 & 58.7 \\
        FCP\cite{park2025foreground} & 941.9 & 5975.1 & 52.5 \\
        VRP-SAM\cite{sun2024vrp} & 472.6 & 5942.3 & 53.9 \\
        UINO-FSS & 93.4 & 5558.0 & 64.5 \\
        \hline
    \end{tabular}
    \label{tab:model-efficiency}
\end{table}

\begin{table}[h]
    \centering
    \small
    \caption{Ablation study results on COCO-20$^i$ under the one-shot setting. Here, we denote BA as the Bottleneck Adapter, PS the Prototype-based Similarity, SVP the Semantic-aware Visual Prompt, MHCM the Mamba-HyperCorrelation Module and CE the Contrastive Enhancement Module.}
    \setlength{\tabcolsep}{3pt} 
    \renewcommand{\arraystretch}{1.3} 
    \begin{tabular*}{\linewidth}{@{\extracolsep{\fill}} c c c c c c @{}}
        \hline
        BA & PS & SVP & MHCM & CE & mIoU (\%) \\
        \hline
        \checkmark & \checkmark & & & & 58.2  \\
        \checkmark & \checkmark & \checkmark & & & 59.6   \\
        \checkmark & & & \checkmark &  & 62.0  \\
        \checkmark & \checkmark & \checkmark & \checkmark &  & 62.3  \\
        \checkmark & \checkmark & \checkmark & \checkmark & \checkmark & \textbf{64.5}  \\
        \hline
    \end{tabular*}
    \label{tab:ablation}
\end{table}

\subsection{Implementation Details}
The lightweight Bottleneck Adapter (BA) is trained on 1\% of the SA-1B dataset \cite{kirillov2023segment} with three stages. In each stage, we train the model with 15 epochs, using an image size of 126$\times$126, 252$\times$252, and 518$\times$518, respectively. In the first two stages, the model is trained on 2 NVIDIA RTX A6000, and in the last stage on 4 RTX A6000 GPUs, with batch size 64. The model is optimized using the AdamW optimizer with an initial learning rate of 1e-3 and a weight decay of 5e-4.

While training for mask decoding and few-shot segmentation, all experiments on the PASCAL-5$^i$ \cite{shaban2017one} and COCO-20$^i$ \cite{nguyen2019feature} datasets were conducted with images of size 518$\times$518, using the frozen encoder from the DINOv2 pre-trained model ViT-B/14 to extract features. To ensure fair comparisons, we followed the same data augmentation and optimization settings as in \cite{peng2023hierarchical}. In particular, images were augmented with scaling, random cropping, rotation, blur, and horizontal flipping, while the learning rate was gradually reduced using a polynomial decay schedule. AdamW was used as the optimizer with an initial learning rate of 0.001 and a weight decay of 1e-2, and the batch size was set to 8 for both datasets.

\subsection{Comparison with State-of-the-Art Methods}
\subsubsection{Few-shot Segmentation Performance} In Tab.~\ref{tab:pascal} and~\ref{tab:coco}, we compare our method with prior works categorized by their backbones. Among foundation model-based approaches, Matcher~\cite{liu2024matcher} and GF-SAM~\cite{zhang2025bridge} are training-free but rely on large pre-trained models, while others, such as VRP-SAM~\cite{sun2024vrp}, FCP~\cite{park2025foreground}, PI-CLIP~\cite{wang2024rethinking}, and SEGIC~\cite{meng2024segic}, undergo supervised training on the target task, i.e., few-shot semantic segmentation, to attain stronger performance. \textbf{Our method gains dramatic improvements over the previous methods with fewer model parameters} (as shown in Tab.~\ref{tab:ablation-complex}) \textbf{on both PASCAL-5$^i$ and COCO-20$^i$ datasets}, thanks to the unified architecture and enhanced meta-visual prompt generator.  

Specifically, on the PASCAL-5$^i$ dataset, our method outperforms PI-CLIP and GF-SAM by 3.8\% and 1.5\% in mIoU, which held the previous memory-free state-of-the-art in the 1-shot and 5-shot settings, respectively. The COCO-20$^i$ dataset presents a more challenging benchmark due to its greater number of categories and higher diversity in object sizes. Even under these conditions, our approach maintains consistently superior performance, exceeding the previous memory-free 1-shot SOTA (VRP-SAM) by 4.1\% and the 5-shot SOTA (GF-SAM) by 2.9\% in mIoU.

From the Tab.~\ref{tab:pascal} and~\ref{tab:coco}, it can be observed that traditional ResNet-based backbones perform significantly worse than methods built upon self-supervised foundation models such as DINOv2 and SAM. Even the best-performing method using ResNet101, namely ABCB, achieves only 72.0\% and 74.9\% in the 1-shot and 5-shot settings, respectively, falling notably behind PI-CLIP and FSSAM. We attribute this gap primarily to the stronger general representation capability of self-supervised pretrained foundation models. Compared with ResNet, which is trained on supervised datasets such as ImageNet, models like DINOv2, SAM, and CLIP are pretrained on large-scale data and thus possess superior semantic alignment and matching capabilities, leading to better performance in downstream tasks.

It is worth noting that our method is fairly compared with state-of-the-art approaches that also leverage foundation-model backbones. As shown in Tab.\ref{tab:pascal} and \ref{tab:coco}, while competing methods adopt heavier or dual-branch architectures—such as VRP-SAM (DINOv2-B + SAM-H), GF-SAM (DINOv2-L + SAM-H), and Matcher (DINOv2-L + SAM-H)—our model uses only a single DINOv2-B backbone. Even with such more efficient setup, our method consistently outperforms these stronger backbones across both PASCAL-\(5^i\) and COCO-\(20^i\) benchmarks. These results confirm that the performance gains stem from our proposed unified framework and modules, rather than from a larger or more powerful backbone. 

In addition, we also compared with methods that are built on SAM2 (an upgraded SAM with memory attention trained on SA-1B and SA-V), such as SANSA~\cite{cuttano2026sansa} and FSSAM~\cite{xu2025unlocking}. As SANSA reports no results on PASCAL-5$^i$ under strict few-shot settings, we compare only on COCO-20$^i$. There, UINO-FSS (DINOv2-B) outperforms SANSA (SAM2-L) by 4.3\% (1-shot) and 5.4\% (5-shot). Compared to FSSAM, our model gains 2.2\% (1-shot) and 0.4\% (5-shot) better mIoU on the challenging COCO-20$^i$ dataset, and gains competitive results on PASCAL-5$^i$. Compared to FSSAM that depends on two foundation models with additional iterative memory refinement and support-calibrated memory attention built on SAM2,  our single-encoder UINO-FSS is considerably more efficient. Given SAM2's superior modeling capacity with memory attention, it is encouraging that UINO-FSS, trained without large-scale video data, competes against this upgraded backbone. This result validates the soundness of our adapter-distillation design and the MVPG with 4D correlation.

\subsubsection{Out-of-Distribution Performance} To evaluate the generalization ability of our model, we conduct one-shot assessments on the FSS-1000. 
Here, FSS-1000 is selected for its task alignment with ours, i.e., segmenting all regions of the support class rather than distinguishing instances.
To assess the generality of our approach, we directly applied a model pretrained on the COCO-20$^i$ base set to FSS-1000 without any additional fine-tuning. This setting is specifically challenging due to the significant domain shifts between the datasets.
The results are presented in Tab.~\ref{tab:zero-shot-rotated}. As shown in the table, our model surpasses SEGIC~\cite{meng2024segic} and PerSAM-F$^{\dagger}$~\cite{zhang2023personalize}, demonstrating outstanding transferability.

\subsubsection{Model Complexity Analysis} Once the distilled model (DINOv2-B \& Bottleneck Adapter) is obtained in the first training stage, it can be frozen and serve as a foundational feature extractor. Based on this, we provide a series of models with incremental trainable parameters: the UINO-FSS-Comp*, UINO-FSS-Comp, and the full UINO-FSS. The UINO-FSS-Comp* retains the same architecture as UINO-FSS but replaces the Mamba-based Dense Prompt Generator (MDPG) with a low-parameter, vanilla dense prompt, which calculates support prototype-based similarity (PS) to the query features.
With its decoder frozen, the total number of trainable parameters is only 0.07M. Based on UINO-FSS-Comp*, the UINO-FSS-Comp variant additionally trains the decoder, yielding a model with 4.1M total trainable parameters. UINO-FSS incorporates MDPG at a cost of merely 0.5M additional parameters, of which the Mamba-based Hyper-Correlation Module (MHCM) accounts for only 0.18M. As shown in Tab.~\ref{tab:ablation-complex}, on COCO-20$^i$, UINO-FSS achieves the best mIoU among all compared methods while maintaining the least total parameters and only a small amount of trainable parameters. Even the UINO-FSS-Comp* can match the performance of GF-SAM with faster inference, and surpass VRP-SAM (ResNet50)\footnote{For VRP-SAM, since the model parameters are only available for its variant with ResNet-50 and SAM-H backbones, here we report the performance (53.9 mIoU) and time complexity of this variant in our comparison.} with +3.3\% mIoU and fewer trainable parameters. Even though our full model introduces limited trainable parameters, its dramatically superior FSS performance, high efficiency, and significantly fewer parameters (only 10\% of GF-SAM), solidly demonstrates its value for FSS.

\subsubsection{Computational Efficiency Analysis}
We compare inference time and peak GPU memory of several vision foundation model-based methods for few-shot segmentation under roughly the same input resolutions (approximately 518×518). As shown in Tab.~\ref{tab:model-efficiency}, our method achieves significantly reduced inference time and also demonstrates clearly lower peak GPU memory.
Due to the limitations of existing FLOPs estimation tools in handling multi-stage pipelines and non-modular operations, some methods cannot be reliably evaluated. Therefore, we use fvcore~\cite{fvcore} to compute FLOPs only for FCP~\cite{park2025foreground} and UINO-FSS-comp, yielding 6236G and 310G, respectively.

\begin{table}[!t]
    \centering
    \footnotesize
    \caption{More ablation studies on HGMB under the one-shot
    setting, where \textsuperscript{\textdagger} denotes the abbreviation for Center-pivot 4D Convolution. We replace HGMB with CP4D Conv, keeping all other modules unchanged. Our HGMB demonstrates significant improvements over CP4D Conv on both COCO-20$^i$ and PASCAL-5$^i$.
     }
    \setlength{\tabcolsep}{1.5pt} 
    \renewcommand{\arraystretch}{1.3} 
    \resizebox{\linewidth}{!}{
    \begin{tabular}{c c c c c c c c}
        \hline
        Dataset & Method & Fold0 & Fold1 & Fold2 & Fold3 & mIoU \\
        \hline
        \multirow{2}{*}{COCO-20$^i$} &  UINO-FSS w/ CP4D Conv\textsuperscript{\textdagger}~\cite{min2021hypercorrelation} & 61.9 & 65.9 & 64.9 & 63.1 & 64.0 \\ & UINO-FSS & \textbf{62.2} & \textbf{66.0} & \textbf{65.5} & \textbf{64.2} & \textbf{64.5} \\
        \hline
        \multirow{2}{*}{PASCAL-5$^i$} &  UINO-FSS w/ CP4D Conv\textsuperscript{\textdagger}~\cite{min2021hypercorrelation} & 79.6 & 82.9 & 77.0 & 79.3 & 79.7 \\ & UINO-FSS & \textbf{80.7} & \textbf{83.2} & \textbf{77.8} & \textbf{80.5} & \textbf{80.6} \\
        \hline
    \end{tabular}
    }
    \label{tab:ablation-MHCM}
\end{table}

\begin{figure*}[t]
\centering
\includegraphics[width=\textwidth]{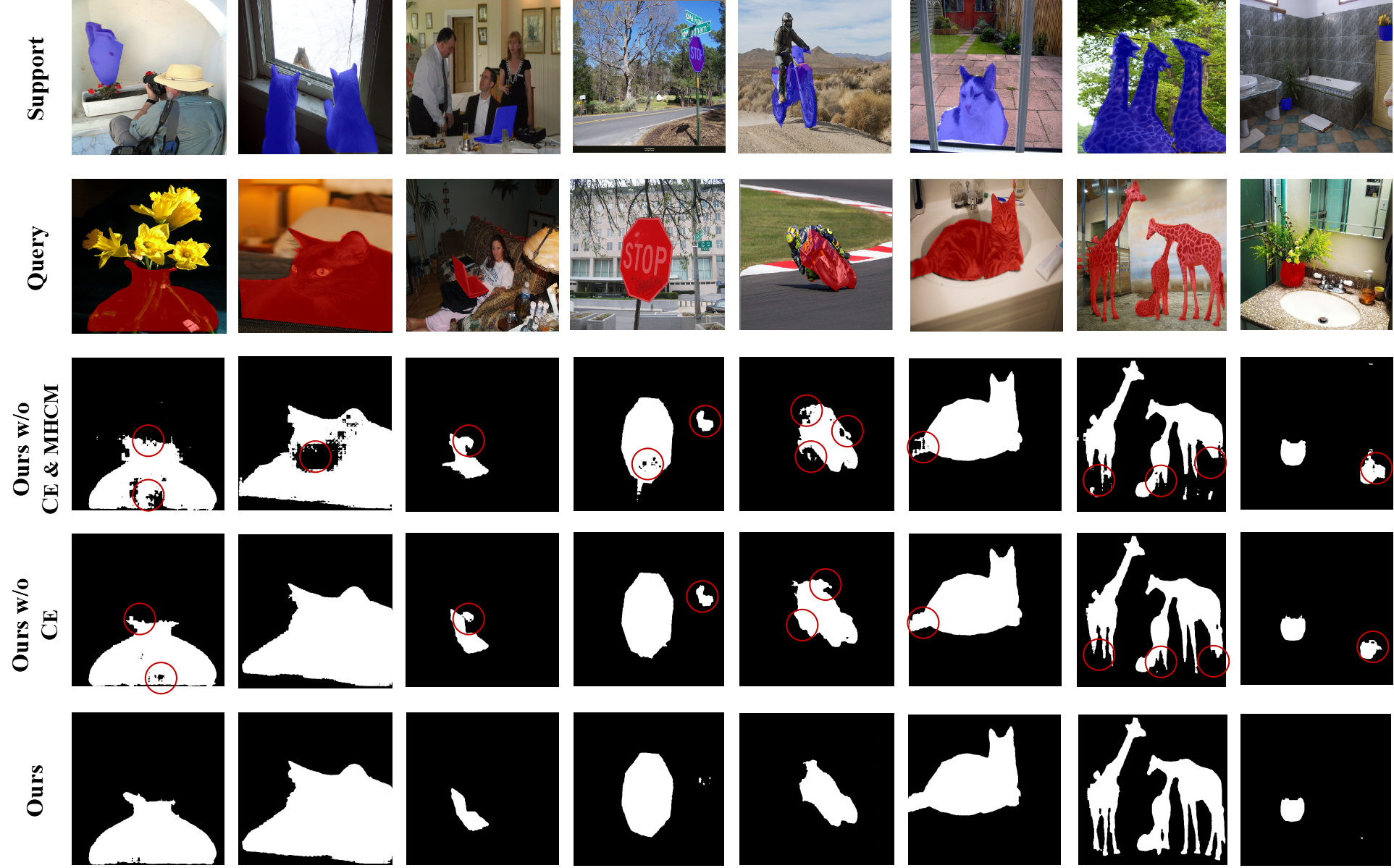}
\caption{\textbf{Qualitative results of UINO-FSS on COCO-20$^i$ under the one-shot setting.
} From top to bottom in each row are the support image with its corresponding mask, the query image with ground-truth annotation, the output of UINO-FSS without CE and MHCM modules, the output of UINO-FSS without CE, and the output of the complete UINO-FSS model. Red circles indicate inaccurate segmentation.}
\label{fig:visualization}
\end{figure*}

\subsection{Ablation Study}
To evaluate our model's effectiveness, we conducted thorough ablation studies on COCO-20$^i$ under one-shot setting. To maintain experimental consistency, we selected DINOv2-Base as the image encoder and trained the mask decoder in all the ablation studies.

\subsubsection{Effectiveness of the  Hierarchical Distillation}
To demonstrate the essential role of hierarchical distillation, here we re-train the compact UINO-FSS with distillation disabled. Specifically, we directly train the bottleneck adapter for few-shot semantic segmentation, with no feature alignment to SAM's encoder. As shown in Tab.~\ref{tab:ablation-distillation}, eliminating distillation results in a sharp \textbf{6.8\%} mIoU decline under the one-shot setting. The resulting 52.8\% mIoU is virtually comparable to SEGIC's mIoU of 53.6\%, that's a baseline training the mask decoder atop a frozen DINOv2 encoder. This outcome strongly affirms the necessity of incorporating knowledge distillation and feature alignment in the framework.  In addition, distillation at single resolution of 126×126 raises CKNNA (top-64) to 0.48 (vs. 0.46 w/o distillation and 0.54 after hierarchical distillation). Using these features in UINO-FSS yields lower mIoU of 63.6\% under the 1-shot setting on COCO-20$^i$, compared to 64.5\% gained by leveraging hierarchical distillation. That proves hierarchical distillation across multiple resolutions provides improved embeddings, and also confirms that SAM's holistic embedding nature--low resolution can already capture its distribution.

\subsubsection{Effectiveness of the Proposed Modules} Unlike prior FSS methods reliant on manually designed sparse prompts (points/boxes), to design a unified framework that eliminates the need for hand-crafted preprocessing, the prompt generator plays an essential role. Theoretically, in the absence of manual coordinate engineering, a sparse prompt cannot provide accurate localization information, thus the dense prompt is critical and unremovable. Based on this rationale, we incrementally add the proposed modules to the framework and have proved their effectiveness and contributions as shown in Tab.~\ref{tab:ablation}. 

Particularly, the \textit{BA+PS} framework serves as the basic baseline. It leverages distilled features from the Bottleneck Adapter (BA) as the image embedding, uses the Prototype-based Similarity (PS) as the dense prompt, and adopts a default, non-conceptual embedding from SAM's original prompt encoder as the sparse prompt. This baseline gets 58.2\% mIoU in one-shot setting, with even 5.5\% performance gains over the strong benchmark of Matcher\cite{liu2024matcher}. This result, again, proves the effectiveness of the unified framework. 
Directly replacing PS with the proposed MHCM can yield a further substantial gain of \textbf{3.8\%} (row 3 vs. row 1 in Tab.~\ref{tab:ablation}). Incrementally adding CE leads to another \textbf{2.2\%} improvement (row 5 vs. row 4).  Note that, the Contrastive Enhancement (CE) is tied with MHCM, since their representations rely on MHCM for modeling. By replacing the default sparse prompt with the semantic-aware visual prompt (SVP) under the naive dense prompt setting of PS, the model~\textit{BA+PS+SVP} achieves a \textbf{1.4\%} performance improvement (comparing row 1 and row 2).
However, the additional \textit{BA+MHCM}-only study presents that PS and SVP contribute marginally when combined with the enhanced dense prompt generator, i.e., the MHCM. This may be because the dense prompts from MHCM already provide effective 2D similarity maps.
To further validate MHCM, we replace it with the Center-pivot 4D convolution (CP4D Conv) module, a convolutional hypercorrelation implementation by~\cite{min2021hypercorrelation}. As shown in Tab.~\ref{tab:ablation-MHCM}, MHCM consistently surpasses CP4D Conv on PASCAL-5$^i$ and COCO-20$^i$. Note that, since the unified framework already established a strong baseline with \textit{BA+SVP+CE}, the additional gains produced by the lightweight MHCM are substantial and non-trivial. Leveraging Mamba's linear complexity, our MHCM alternates VSS and convolution modules to efficiently capture global dependencies and propagate information across the entire 4D hypercorrelation volume via patch-wise aggregation and residual fusion. Unlike traditional methods that stack convolutions on deformed 2D maps, our approach expands the receptive field more effectively and captures long-range matching relationships with higher accuracy. This could be why the proposed MHCM yields better results.

In addition, we visualize the segmentation results in Fig.~\ref{fig:visualization}. We can see that the framework without CE and MHCM can already produce basically accurate segmentation, but the predicted masks involve holes, missing details and wrongly segmented backgrounds. The MHCM that models neighboring correlations in 4D space effectively blends the segmentation with enhanced details. The foreground/background contrastive enhancement in 4D correlation volume further removes wrongly segmented regions on backgrounds.

\begin{table}[!t]
    \centering
    \footnotesize
    \caption{ More studies on cross-model distillation from coarse to fine. }
    \setlength{\tabcolsep}{3pt} 
    \renewcommand{\arraystretch}{1.3} 
    \resizebox{\linewidth}{!}{
    \begin{tabular}{c c c c c c c}
        \hline
        Method & Fold0 & Fold1 & Fold2 & Fold3 & mIoU \\
        \hline
        UINO-FSS-Comp w/o distillation & \underline{}{50.5} & \underline{}{55.2} & \underline{}{53.5} & \underline{}{52.1} & \underline{}{52.8} \\
        UINO-FSS-Comp & \textbf{55.5} & \textbf{63.8} & \textbf{59.4} & \textbf{59.5} & \textbf{59.6} \\
        \hline
    \end{tabular}
    }
    \label{tab:ablation-distillation}
\end{table}

\begin{table}[!t]
    \centering
    \footnotesize 
    \caption{Performance comparison of our compact model with different DINOv2 backbones (small/base/large) on COCO-20$^i$ under 1-shot setting. }
    \setlength{\tabcolsep}{3pt} 
    \renewcommand{\arraystretch}{1.3} 
    \begin{tabularx}{\linewidth}{c *{5}{>{\centering\arraybackslash}X} c}
        \hline
        Method & Fold0 & Fold1 & Fold2 & Fold3 & mIoU \\
        \hline
         UINO-FSS-Comp w/ DINOv2-Small & 49.2 & 58.5 & 55.3 & 54.3 & 54.3 \\
         UINO-FSS-Comp w/ DINOv2-Base & \underline{}{55.5} & \textbf{}{63.8} & \underline{}{59.4} & \underline{}{59.5} & \underline{}{59.6} \\
         UINO-FSS-Comp w/ DINOv2-Large &\textbf{}{57.6} & \textbf{}{63.8} & \textbf{}{61.6} & \textbf{}{59.9} & \textbf{}{60.7} \\
        \hline
    \end{tabularx}
    \label{tab:ablation-dinov2-family}
\end{table}

\subsubsection{Generalization on DINOv2 Variations}
For generalization, we inspect the self-similarity map (Eq.~\ref{eq:SSM}) produced by DINOv2-Small and DINOv2-Large, and select the 2nd and 4th layer respectively for distillation. The mask decoder is then trained on the distilled image features for few-shot semantic segmentation. Tab.~\ref{tab:ablation-dinov2-family} confirms that stronger backbones yield steadily higher mIoU, with the largest gain occurring when switching from DINOv2-Small to DINOv2-Base. Upgrading to DINOv2-Large still helps, but the improvement is modest; This could be attributed to the fact that SAM-H has already transferred most of its knowledge to DINOv2-Base through the bottleneck adapter, leaving limited extra information for the larger model to absorb. Nevertheless, the 60.7\% mIoU still exceeds the 58.7\% achieved by GF-SAM, that is built on both DINOv2-Large and SAM-H, attesting to the added value of our design. The pretrained models in Tab.~\ref{tab:ablation-dinov2-family} can also serve as new foundation models for few-shot semantic segmentation.

\section{Conclusion}
In summary, we introduce a novel unified framework for few-shot semantic segmentation and its compact variations built upon DINOv2 families. Specifically, in this work, we utilize vision foundation models for few-shot segmentation, finding that DINOv2 features offer strong discriminability across classes. This enables the generation of accurate prior information to guide the segmentation process. To enhance efficiency and accelerate inference, we design a lightweight bottleneck adapter on the third-layer features of DINOv2’s image encoder, which learns from SAM’s encoder through coarse-to-fine distillation. Additionally, we propose a meta-visual prompt generator that leverages the capabilities of large-scale pre-trained models and uses the mask decoder initialized from SAM for complete decoding. For visual prompting, both the semantic-aware visual prompts and Mamba-based dense prompting contribute to the final improvements. Extensive experiments demonstrate the effectiveness of our model, and future work will explore further improvements in accuracy and generalization.

 \bibliographystyle{IEEEtran}
\bibliography{IEEEabrv,./paper}

\begin{thebibliography}{10}
\providecommand{\url}[1]{#1}
\csname url@samestyle\endcsname
\providecommand{\newblock}{\relax}
\providecommand{\bibinfo}[2]{#2}
\providecommand{\BIBentrySTDinterwordspacing}{\spaceskip=0pt\relax}
\providecommand{\BIBentryALTinterwordstretchfactor}{4}
\providecommand{\BIBentryALTinterwordspacing}{\spaceskip=\fontdimen2\font plus
\BIBentryALTinterwordstretchfactor\fontdimen3\font minus \fontdimen4\font\relax}
\providecommand{\BIBforeignlanguage}[2]{{%
\expandafter\ifx\csname l@#1\endcsname\relax
\typeout{** WARNING: IEEEtran.bst: No hyphenation pattern has been}%
\typeout{** loaded for the language `#1'. Using the pattern for}%
\typeout{** the default language instead.}%
\else
\language=\csname l@#1\endcsname
\fi
#2}}
\providecommand{\BIBdecl}{\relax}
\BIBdecl

\bibitem{liu2024matcher}
Y.~Liu, M.~Zhu, H.~Li, H.~Chen, X.~Wang, and C.~Shen, ``Matcher: Segment anything with one shot using all-purpose feature matching,'' in \emph{ICLR}, 2024.

\bibitem{zhang2025bridge}
A.~Zhang, G.~Gao, J.~Jiao, C.~Liu, and Y.~Wei, ``Bridge the points: Graph-based few-shot segment anything semantically,'' \emph{Advances in Neural Information Processing Systems}, 2024.

\bibitem{sun2024vrp}
Y.~Sun, J.~Chen, S.~Zhang, X.~Zhang, Q.~Chen, G.~Zhang, E.~Ding, J.~Wang, and Z.~Li, ``Vrp-sam: Sam with visual reference prompt,'' in \emph{Proceedings of the IEEE/CVF Conference on Computer Vision and Pattern Recognition}, 2024, pp. 23\,565--23\,574.

\bibitem{snell2017prototypical}
J.~Snell, K.~Swersky, and R.~Zemel, ``Prototypical networks for few-shot learning,'' \emph{Advances in neural information processing systems}, vol.~30, 2017.

\bibitem{shaban2017one}
A.~Shaban, S.~Bansal, Z.~Liu, I.~Essa, and B.~Boots, ``One-shot learning for semantic segmentation,'' \emph{arXiv preprint arXiv:1709.03410}, 2017.

\bibitem{wang2019panet}
K.~Wang, J.~H. Liew, Y.~Zou, D.~Zhou, and J.~Feng, ``Panet: Few-shot image semantic segmentation with prototype alignment,'' in \emph{proceedings of the IEEE/CVF international conference on computer vision}, 2019, pp. 9197--9206.

\bibitem{liu2020part}
Y.~Liu, X.~Zhang, S.~Zhang, and X.~He, ``Part-aware prototype network for few-shot semantic segmentation,'' in \emph{Computer Vision--ECCV 2020: 16th European Conference, Glasgow, UK, August 23--28, 2020, Proceedings, Part IX 16}.\hskip 1em plus 0.5em minus 0.4em\relax Springer, 2020, pp. 142--158.

\bibitem{yang2020prototype}
B.~Yang, C.~Liu, B.~Li, J.~Jiao, and Q.~Ye, ``Prototype mixture models for few-shot semantic segmentation,'' in \emph{Computer Vision--ECCV 2020: 16th European Conference, Glasgow, UK, August 23--28, 2020, Proceedings, Part VIII 16}.\hskip 1em plus 0.5em minus 0.4em\relax Springer, 2020, pp. 763--778.

\bibitem{latent-mining}
L.~Yang, W.~Zhuo, L.~Qi, Y.~Shi, and Y.~Gao, ``Mining latent classes for few-shot segmentation,'' in \emph{Proceedings of the IEEE/CVF international conference on computer vision}, 2021, pp. 8721--8730.

\bibitem{zhang2021self}
B.~Zhang, J.~Xiao, and T.~Qin, ``Self-guided and cross-guided learning for few-shot segmentation,'' in \emph{Proceedings of the IEEE/CVF conference on computer vision and pattern recognition}, 2021, pp. 8312--8321.

\bibitem{zhang2020sg}
X.~Zhang, Y.~Wei, Y.~Yang, and T.~S. Huang, ``Sg-one: Similarity guidance network for one-shot semantic segmentation,'' \emph{IEEE transactions on cybernetics}, vol.~50, no.~9, pp. 3855--3865, 2020.

\bibitem{gao2022drnet}
G.~Gao, Z.~Fang, C.~Han, Y.~Wei, C.~H. Liu, and S.~Yan, ``Drnet: Double recalibration network for few-shot semantic segmentation,'' \emph{IEEE Transactions on Image Processing}, vol.~31, pp. 6733--6746, 2022.

\bibitem{chen2024dual}
Y.~Chen, R.~Jiang, Y.~Zheng, B.~Sheng, Z.-X. Yang, and E.~Wu, ``Dual branch multi-level semantic learning for few-shot segmentation,'' \emph{IEEE Transactions on Image Processing}, vol.~33, pp. 1432--1447, 2024.

\bibitem{zhang2021few}
G.~Zhang, G.~Kang, Y.~Yang, and Y.~Wei, ``Few-shot segmentation via cycle-consistent transformer,'' \emph{Advances in Neural Information Processing Systems}, vol.~34, pp. 21\,984--21\,996, 2021.

\bibitem{iqbal2022msanet}
E.~Iqbal, S.~Safarov, and S.~Bang, ``Msanet: Multi-similarity and attention guidance for boosting few-shot segmentation,'' \emph{arXiv preprint arXiv:2206.09667}, 2022.

\bibitem{peng2023hierarchical}
B.~Peng, Z.~Tian, X.~Wu, C.~Wang, S.~Liu, J.~Su, and J.~Jia, ``Hierarchical dense correlation distillation for few-shot segmentation,'' in \emph{Proceedings of the IEEE/CVF Conference on Computer Vision and Pattern Recognition}, 2023, pp. 23\,641--23\,651.

\bibitem{min2021hypercorrelation}
J.~Min, D.~Kang, and M.~Cho, ``Hypercorrelation squeeze for few-shot segmentation,'' in \emph{Proceedings of the IEEE/CVF international conference on computer vision}, 2021, pp. 6941--6952.

\bibitem{hong2022cost}
S.~Hong, S.~Cho, J.~Nam, S.~Lin, and S.~Kim, ``Cost aggregation with 4d convolutional swin transformer for few-shot segmentation,'' in \emph{European Conference on Computer Vision}.\hskip 1em plus 0.5em minus 0.4em\relax Springer, 2022, pp. 108--126.

\bibitem{xu2023self}
Q.~Xu, W.~Zhao, G.~Lin, and C.~Long, ``Self-calibrated cross attention network for few-shot segmentation,'' in \emph{Proceedings of the IEEE/CVF international conference on computer vision}, 2023, pp. 655--665.

\bibitem{caron2021emerging}
M.~Caron, H.~Touvron, I.~Misra, H.~J{\'e}gou, J.~Mairal, P.~Bojanowski, and A.~Joulin, ``Emerging properties in self-supervised vision transformers,'' in \emph{Proceedings of the IEEE/CVF international conference on computer vision}, 2021, pp. 9650--9660.

\bibitem{oquab2024dinov2}
M.~Oquab, T.~Darcet, T.~Moutakanni, H.~Vo, M.~Szafraniec, V.~Khalidov, P.~Fernandez, D.~Haziza, F.~Massa, A.~El-Nouby \emph{et~al.}, ``Dinov2: Learning robust visual features without supervision,'' \emph{Transactions on Machine Learning Research Journal}, 2024.

\bibitem{radford2021learning}
A.~Radford, J.~W. Kim, C.~Hallacy, A.~Ramesh, G.~Goh, S.~Agarwal, G.~Sastry, A.~Askell, P.~Mishkin, J.~Clark \emph{et~al.}, ``Learning transferable visual models from natural language supervision,'' in \emph{International conference on machine learning}.\hskip 1em plus 0.5em minus 0.4em\relax PMLR, 2021, pp. 8748--8763.

\bibitem{kirillov2023segment}
A.~Kirillov, E.~Mintun, N.~Ravi, H.~Mao, C.~Rolland, L.~Gustafson, T.~Xiao, S.~Whitehead, A.~C. Berg, W.-Y. Lo \emph{et~al.}, ``Segment anything,'' in \emph{Proceedings of the IEEE/CVF international conference on computer vision}, 2023, pp. 4015--4026.

\bibitem{he2024apseg}
W.~He, Y.~Zhang, W.~Zhuo, L.~Shen, J.~Yang, S.~Deng, and L.~Sun, ``Apseg: auto-prompt network for cross-domain few-shot semantic segmentation,'' in \emph{Proceedings of the IEEE/CVF Conference on Computer Vision and Pattern Recognition}, 2024, pp. 23\,762--23\,772.

\bibitem{zhang2023personalize}
R.~Zhang, Z.~Jiang, Z.~Guo, S.~Yan, J.~Pan, X.~Ma, H.~Dong, P.~Gao, and H.~Li, ``Personalize segment anything model with one shot,'' \emph{arXiv preprint arXiv:2305.03048}, 2023.

\bibitem{meng2024segic}
L.~Meng, S.~Lan, H.~Li, J.~M. Alvarez, Z.~Wu, and Y.-G. Jiang, ``Segic: Unleashing the emergent correspondence for in-context segmentation,'' in \emph{European Conference on Computer Vision}.\hskip 1em plus 0.5em minus 0.4em\relax Springer, 2024, pp. 203--220.

\bibitem{nguyen2019feature}
K.~Nguyen and S.~Todorovic, ``Feature weighting and boosting for few-shot segmentation,'' in \emph{Proceedings of the IEEE/CVF international conference on computer vision}, 2019, pp. 622--631.

\bibitem{xu2025unlocking}
Q.~Xu, L.~Zhu, X.~Liu, G.~Lin, C.~Long, Z.~Li, and R.~Zhao, ``Unlocking the power of sam 2 for few-shot segmentation,'' \emph{arXiv preprint arXiv:2505.14100}, 2025.

\bibitem{ronneberger2015u}
O.~Ronneberger, P.~Fischer, and T.~Brox, ``U-net: Convolutional networks for biomedical image segmentation,'' in \emph{International Conference on Medical image computing and computer-assisted intervention}.\hskip 1em plus 0.5em minus 0.4em\relax Springer, 2015, pp. 234--241.

\bibitem{badrinarayanan2017segnet}
V.~Badrinarayanan, A.~Kendall, and R.~Cipolla, ``Segnet: A deep convolutional encoder-decoder architecture for image segmentation,'' \emph{IEEE transactions on pattern analysis and machine intelligence}, vol.~39, no.~12, pp. 2481--2495, 2017.

\bibitem{yu2015multi}
F.~Yu and V.~Koltun, ``Multi-scale context aggregation by dilated convolutions,'' \emph{arXiv preprint arXiv:1511.07122}, 2015.

\bibitem{chen2017rethinking}
L.-C. Chen, G.~Papandreou, F.~Schroff, and H.~Adam, ``Rethinking atrous convolution for semantic image segmentation,'' \emph{arXiv preprint arXiv:1706.05587}, 2017.

\bibitem{chen2018encoder}
L.-C. Chen, Y.~Zhu, G.~Papandreou, F.~Schroff, and H.~Adam, ``Encoder-decoder with atrous separable convolution for semantic image segmentation,'' in \emph{Proceedings of the European conference on computer vision (ECCV)}, 2018, pp. 801--818.

\bibitem{zhao2017pyramid}
H.~Zhao, J.~Shi, X.~Qi, X.~Wang, and J.~Jia, ``Pyramid scene parsing network,'' in \emph{Proceedings of the IEEE conference on computer vision and pattern recognition}, 2017, pp. 2881--2890.

\bibitem{sun2019high}
K.~Sun, Y.~Zhao, B.~Jiang, T.~Cheng, B.~Xiao, D.~Liu, Y.~Mu, X.~Wang, W.~Liu, and J.~Wang, ``High-resolution representations for labeling pixels and regions,'' \emph{arXiv preprint arXiv:1904.04514}, 2019.

\bibitem{qin2024pyramid}
Z.~Qin, J.~Liu, X.~Zhang, M.~Tian, A.~Zhou, S.~Yi, and H.~Li, ``Pyramid fusion transformer for semantic segmentation,'' \emph{IEEE Transactions on Multimedia}, vol.~26, pp. 9630--9643, 2024.

\bibitem{fu2019dual}
J.~Fu, J.~Liu, H.~Tian, Y.~Li, Y.~Bao, Z.~Fang, and H.~Lu, ``Dual attention network for scene segmentation,'' in \emph{Proceedings of the IEEE/CVF conference on computer vision and pattern recognition}, 2019, pp. 3146--3154.

\bibitem{huang2019ccnet}
Z.~Huang, X.~Wang, L.~Huang, C.~Huang, Y.~Wei, and W.~Liu, ``Ccnet: Criss-cross attention for semantic segmentation,'' in \emph{Proceedings of the IEEE/CVF international conference on computer vision}, 2019, pp. 603--612.

\bibitem{cheng2022masked}
B.~Cheng, I.~Misra, A.~G. Schwing, A.~Kirillov, and R.~Girdhar, ``Masked-attention mask transformer for universal image segmentation,'' in \emph{Proceedings of the IEEE/CVF conference on computer vision and pattern recognition}, 2022, pp. 1290--1299.

\bibitem{xie2021segformer}
E.~Xie, W.~Wang, Z.~Yu, A.~Anandkumar, J.~M. Alvarez, and P.~Luo, ``Segformer: Simple and efficient design for semantic segmentation with transformers,'' \emph{Advances in neural information processing systems}, vol.~34, pp. 12\,077--12\,090, 2021.

\bibitem{wen2024rethinking}
Q.~Wen and C.-G. Li, ``Rethinking decoders for transformer-based semantic segmentation: A compression perspective,'' \emph{Advances in Neural Information Processing Systems}, vol.~37, pp. 49\,806--49\,833, 2024.

\bibitem{wang2023pcnet}
J.-Y. Wang, S.-K. Liu, S.-C. Guo, C.-Y. Jiang, and W.-M. Zheng, ``Pcnet: Leveraging prototype complementarity to improve prototype affinity for few-shot segmentation,'' \emph{Electronics}, vol.~13, no.~1, p. 142, 2023.

\bibitem{liu2021harmonic}
B.~Liu, J.~Jiao, and Q.~Ye, ``Harmonic feature activation for few-shot semantic segmentation,'' \emph{IEEE Transactions on Image Processing}, vol.~30, pp. 3142--3153, 2021.

\bibitem{zhuge2021deep}
Y.~Zhuge and C.~Shen, ``Deep reasoning network for few-shot semantic segmentation,'' in \emph{Proceedings of the 29th ACM International Conference on Multimedia}, 2021, pp. 5344--5352.

\bibitem{moon2023msi}
S.~Moon, S.~S. Sohn, H.~Zhou, S.~Yoon, V.~Pavlovic, M.~H. Khan, and M.~Kapadia, ``Msi: Maximize support-set information for few-shot segmentation,'' in \emph{Proceedings of the IEEE/CVF International Conference on Computer Vision}, 2023, pp. 19\,266--19\,276.

\bibitem{lu2021simpler}
Z.~Lu, S.~He, X.~Zhu, L.~Zhang, Y.-Z. Song, and T.~Xiang, ``Simpler is better: Few-shot semantic segmentation with classifier weight transformer,'' in \emph{Proceedings of the IEEE/CVF International Conference on Computer Vision}, 2021, pp. 8741--8750.

\bibitem{shi2022dense}
X.~Shi, D.~Wei, Y.~Zhang, D.~Lu, M.~Ning, J.~Chen, K.~Ma, and Y.~Zheng, ``Dense cross-query-and-support attention weighted mask aggregation for few-shot segmentation,'' in \emph{European Conference on Computer Vision}.\hskip 1em plus 0.5em minus 0.4em\relax Springer, 2022, pp. 151--168.

\bibitem{chang2024drnet}
Z.~Chang, X.~Gao, N.~Li, H.~Zhou, and Y.~Lu, ``Drnet: Disentanglement and recombination network for few-shot semantic segmentation,'' \emph{IEEE Transactions on Circuits and Systems for Video Technology}, vol.~34, no.~7, pp. 5560--5574, 2024.

\bibitem{luo2023pfenet++}
X.~Luo, Z.~Tian, T.~Zhang, B.~Yu, Y.~Y. Tang, and J.~Jia, ``Pfenet++: Boosting few-shot semantic segmentation with the noise-filtered context-aware prior mask,'' \emph{IEEE Transactions on Pattern Analysis and Machine Intelligence}, vol.~46, no.~2, pp. 1273--1289, 2023.

\bibitem{lu2023prediction}
Z.~Lu, S.~He, D.~Li, Y.-Z. Song, and T.~Xiang, ``Prediction calibration for generalized few-shot semantic segmentation,'' \emph{IEEE transactions on image processing}, vol.~32, pp. 3311--3323, 2023.

\bibitem{dosovitskiy2020image}
A.~Dosovitskiy, L.~Beyer, A.~Kolesnikov, D.~Weissenborn, X.~Zhai, T.~Unterthiner, M.~Dehghani, M.~Minderer, G.~Heigold, S.~Gelly \emph{et~al.}, ``An image is worth 16x16 words: Transformers for image recognition at scale,'' \emph{arXiv preprint arXiv:2010.11929}, 2020.

\bibitem{chen2024visual}
S.~Chen, F.~Meng, R.~Zhang, H.~Qiu, H.~Li, Q.~Wu, and L.~Xu, ``Visual and textual prior guided mask assemble for few-shot segmentation and beyond,'' \emph{IEEE Transactions on Multimedia}, vol.~26, pp. 7197--7209, 2024.

\bibitem{wang2024rethinking}
J.~Wang, B.~Zhang, J.~Pang, H.~Chen, and W.~Liu, ``Rethinking prior information generation with clip for few-shot segmentation,'' in \emph{Proceedings of the IEEE/CVF Conference on Computer Vision and Pattern Recognition}, 2024, pp. 3941--3951.

\bibitem{ravi2025sam}
N.~Ravi, V.~Gabeur, Y.-T. Hu, R.~Hu, C.~Ryali, T.~Ma, H.~Khedr, R.~R{\"a}dle, C.~Rolland, L.~Gustafson \emph{et~al.}, ``Sam 2: Segment anything in images and videos,'' in \emph{International Conference on Learning Representations}, vol. 2025, 2025, pp. 28\,085--28\,128.

\bibitem{cuttano2026sansa}
C.~Cuttano, G.~Trivigno, G.~Averta, and C.~Masone, ``Sansa: Unleashing the hidden semantics in sam2 for few-shot segmentation,'' \emph{Advances in Neural Information Processing Systems}, 2025.

\bibitem{heinrich2025radiov2}
G.~Heinrich, M.~Ranzinger, H.~Yin, Y.~Lu, J.~Kautz, A.~Tao, B.~Catanzaro, and P.~Molchanov, ``Radiov2. 5: Improved baselines for agglomerative vision foundation models,'' in \emph{Proceedings of the Computer Vision and Pattern Recognition Conference}, 2025, pp. 22\,487--22\,497.

\bibitem{gu2023mamba}
A.~Gu and T.~Dao, ``Mamba: Linear-time sequence modeling with selective state spaces,'' \emph{arXiv preprint arXiv:2312.00752}, 2023.

\bibitem{xu2024hybrid}
Q.~Xu, X.~Liu, L.~Zhu, G.~Lin, C.~Long, Z.~Li, and R.~Zhao, ``Hybrid mamba for few-shot segmentation,'' \emph{Advances in Neural Information Processing Systems}, vol.~37, pp. 73\,858--73\,883, 2024.

\bibitem{huang2024localmamba}
T.~Huang, X.~Pei, S.~You, F.~Wang, C.~Qian, and C.~Xu, ``Localmamba: Visual state space model with windowed selective scan,'' in \emph{European conference on computer vision}.\hskip 1em plus 0.5em minus 0.4em\relax Springer, 2024, pp. 12--22.

\bibitem{xiao2025spatial}
C.~Xiao, M.~Li, D.~Meng, L.~Zhang \emph{et~al.}, ``Spatial-mamba: Effective visual state space models via structure-aware state fusion,'' in \emph{International Conference on Learning Representations}, vol. 2025, 2025, pp. 73\,777--73\,795.

\bibitem{liu2025defmamba}
L.~Liu, M.~Zhang, J.~Yin, T.~Liu, W.~Ji, Y.~Piao, and H.~Lu, ``Defmamba: Deformable visual state space model,'' in \emph{Proceedings of the Computer Vision and Pattern Recognition Conference}, 2025, pp. 8838--8847.

\bibitem{shaker2025groupmamba}
A.~Shaker, S.~T. Wasim, S.~Khan, J.~Gall, and F.~S. Khan, ``Groupmamba: Efficient group-based visual state space model,'' in \emph{Proceedings of the Computer Vision and Pattern Recognition Conference}, 2025, pp. 14\,912--14\,922.

\bibitem{cheng2026mambapanoptic}
Q.~Cheng, D.~Bertolini, W.~Zhang, D.~Wang, N.~Zeller, and D.~Cremers, ``Mambapanoptic: A vision mamba-based structured state space framework for panoptic segmentation,'' \emph{arXiv preprint arXiv:2605.12640}, 2026.

\bibitem{liu2024vmamba}
Y.~Liu, Y.~Tian, Y.~Zhao, H.~Yu, L.~Xie, Y.~Wang, Q.~Ye, J.~Jiao, and Y.~Liu, ``Vmamba: Visual state space model,'' \emph{Advances in neural information processing systems}, vol.~37, pp. 103\,031--103\,063, 2024.

\bibitem{cknna}
M.~Huh, B.~Cheung, T.~Wang, and P.~Isola, ``Position: The platonic representation hypothesis,'' in \emph{Forty-first International Conference on Machine Learning}, 2024.

\bibitem{hsic}
A.~Gretton, O.~Bousquet, A.~Smola, and B.~Sch{\"o}lkopf, ``Measuring statistical dependence with hilbert-schmidt norms,'' in \emph{International conference on algorithmic learning theory}.\hskip 1em plus 0.5em minus 0.4em\relax Springer, 2005, pp. 63--77.

\bibitem{everingham2010pascal}
M.~Everingham, L.~Van~Gool, C.~K. Williams, J.~Winn, and A.~Zisserman, ``The pascal visual object classes (voc) challenge,'' \emph{International journal of computer vision}, vol.~88, pp. 303--338, 2010.

\bibitem{hariharan2011semantic}
B.~Hariharan, P.~Arbel{\'a}ez, L.~Bourdev, S.~Maji, and J.~Malik, ``Semantic contours from inverse detectors,'' in \emph{2011 international conference on computer vision}.\hskip 1em plus 0.5em minus 0.4em\relax IEEE, 2011, pp. 991--998.

\bibitem{lin2014microsoft}
T.-Y. Lin, M.~Maire, S.~Belongie, J.~Hays, P.~Perona, D.~Ramanan, P.~Doll{\'a}r, and C.~L. Zitnick, ``Microsoft coco: Common objects in context,'' in \emph{Computer vision--ECCV 2014: 13th European conference, zurich, Switzerland, September 6-12, 2014, proceedings, part v 13}.\hskip 1em plus 0.5em minus 0.4em\relax Springer, 2014, pp. 740--755.

\bibitem{li2020fss}
X.~Li, T.~Wei, Y.~P. Chen, Y.-W. Tai, and C.-K. Tang, ``Fss-1000: A 1000-class dataset for few-shot segmentation,'' in \emph{Proceedings of the IEEE/CVF conference on computer vision and pattern recognition}, 2020, pp. 2869--2878.

\bibitem{lang2022learning}
C.~Lang, G.~Cheng, B.~Tu, and J.~Han, ``Learning what not to segment: A new perspective on few-shot segmentation,'' in \emph{Proceedings of the IEEE/CVF conference on computer vision and pattern recognition}, 2022, pp. 8057--8067.

\bibitem{xu2024eliminating}
Q.~Xu, G.~Lin, C.~C. Loy, C.~Long, Z.~Li, and R.~Zhao, ``Eliminating feature ambiguity for few-shot segmentation,'' in \emph{European Conference on Computer Vision}.\hskip 1em plus 0.5em minus 0.4em\relax Springer, 2024, pp. 416--433.

\bibitem{wen2025object}
C.~Wen, Y.~Zhang, J.~Fan, H.~Zhu, X.-S. Wei, Y.~Wang, Z.~Kou, and S.~Sun, ``Object-level correlation for few-shot segmentation,'' in \emph{Proceedings of the IEEE/CVF International Conference on Computer Vision}, 2025, pp. 23\,689--23\,699.

\bibitem{zhu2024addressing}
L.~Zhu, T.~Chen, J.~Yin, S.~See, and J.~Liu, ``Addressing background context bias in few-shot segmentation through iterative modulation,'' in \emph{Proceedings of the IEEE/CVF conference on computer vision and pattern recognition}, 2024, pp. 3370--3379.

\bibitem{park2025foreground}
S.~Park, S.~Lee, H.~S. Seong, J.~Yoo, and J.-P. Heo, ``Foreground-covering prototype generation and matching for sam-aided few-shot segmentation,'' in \emph{Proceedings of the AAAI Conference on Artificial Intelligence}, vol.~39, no.~6, 2025, pp. 6425--6433.

\bibitem{wang2023images}
X.~Wang, W.~Wang, Y.~Cao, C.~Shen, and T.~Huang, ``Images speak in images: A generalist painter for in-context visual learning,'' in \emph{Proceedings of the IEEE/CVF Conference on Computer Vision and Pattern Recognition}, 2023, pp. 6830--6839.

\bibitem{wang2023seggpt}
X.~Wang, X.~Zhang, Y.~Cao, W.~Wang, C.~Shen, and T.~Huang, ``Seggpt: Towards segmenting everything in context,'' in \emph{Proceedings of the IEEE/CVF International Conference on Computer Vision}, 2023, pp. 1130--1140.

\bibitem{fvcore}
{Facebook AI Research}, \url{https://github.com/facebookresearch/fvcore}.

\end{thebibliography}

\vfill

\end{document}